\documentclass{article}
\usepackage{amsmath}
\usepackage{microtype}
\usepackage{graphicx}
\usepackage{subcaption}
\usepackage{booktabs} 
\usepackage{tabularx}
\usepackage{colortbl} 
\usepackage{multirow}
\usepackage{graphicx}
\usepackage{hyperref}
\usepackage{threeparttable}
\usepackage{booktabs}

\usepackage[preprint]{icml2026}

\usepackage{amsmath}
\usepackage{amssymb}
\usepackage{mathtools}
\usepackage{amsthm}
\usepackage{tcolorbox}
\tcbuselibrary{breakable,listings}
\usepackage{listings}
\usepackage{xcolor}

\usepackage[capitalize,noabbrev]{cleveref}

\theoremstyle{plain}

\theoremstyle{definition}

\theoremstyle{remark}

\usepackage[textsize=tiny]{todonotes}

\icmltitlerunning{Beyond Rating: A Comprehensive Evaluation and Benchmark for AI Reviews}

\begin{document}

\twocolumn[
  \icmltitle{Beyond Rating: A Comprehensive Evaluation and Benchmark for AI Reviews}



  \icmlsetsymbol{equal}{*}
  \icmlsetsymbol{corr}{\dag}

  \begin{icmlauthorlist}
    \icmlauthor{Bowen Li}{equal,fd}
    \icmlauthor{Haochen Ma}{equal,fd}
    \icmlauthor{Yuxin Wang}{fd}
    \icmlauthor{Jie Yang}{fd}
    \icmlauthor{Yining Zheng}{corr,fd}
    \icmlauthor{Xinchi Chen}{fd}
    \icmlauthor{Xuanjing Huang}{fd}
    \icmlauthor{Xipeng Qiu}{corr,fd,sii}

  \end{icmlauthorlist}

  \icmlaffiliation{fd}{College of Computer Science and Artificial Intelligence, Fudan University, Shanghai, China}

  \icmlaffiliation{sii}{Shanghai Innovation Institute, Shanghai, China}

  \icmlcorrespondingauthor{Yining Zheng}{ynzheng@fudan.edu.cn}
  \icmlcorrespondingauthor{Xipeng Qiu}{xpqiu@fudan.edu.cn}

  \icmlkeywords{Machine Learning, ICML}

  \vskip 0.3in
]



\printAffiliationsAndNotice{}  

\newcommand{\yxwang}[1]{{\color{blue} [yxwang: ``#1'']}}

\begin{abstract}

The rapid adoption of Large Language Models (LLMs) has spurred interest in automated peer review; however, progress is currently stifled by benchmarks that treat reviewing primarily as a rating prediction task. We argue that the utility of a review lies in its textual justification—its arguments, questions, and critique—rather than a scalar score. To address this, we introduce \textbf{Beyond Rating}, a holistic evaluation framework that assesses AI reviewers across five dimensions: Content Faithfulness, Argumentative Alignment, Focus Consistency, Question Constructiveness, and AI-Likelihood. Notably, we propose a ``Max-Recall'' strategy to accommodate valid expert disagreement and introduce a curated dataset of paper with high-confidence reviews, rigorously filtered to remove procedural noise. Extensive experiments demonstrate that while traditional n-gram metrics fail to reflect human preferences, our proposed text-centric metrics—particularly the recall of weakness arguments—correlate strongly with rating accuracy. These findings establish that aligning AI critique focus with human experts is a prerequisite for reliable automated scoring, offering a robust standard for future research.

\end{abstract}

\section{Introduction}
\label{sec:intro}

\begin{figure}[htbp]
    \centering
    \begin{subfigure}[t]{0.4\textwidth}
        \centering
        \includegraphics[width=\linewidth]{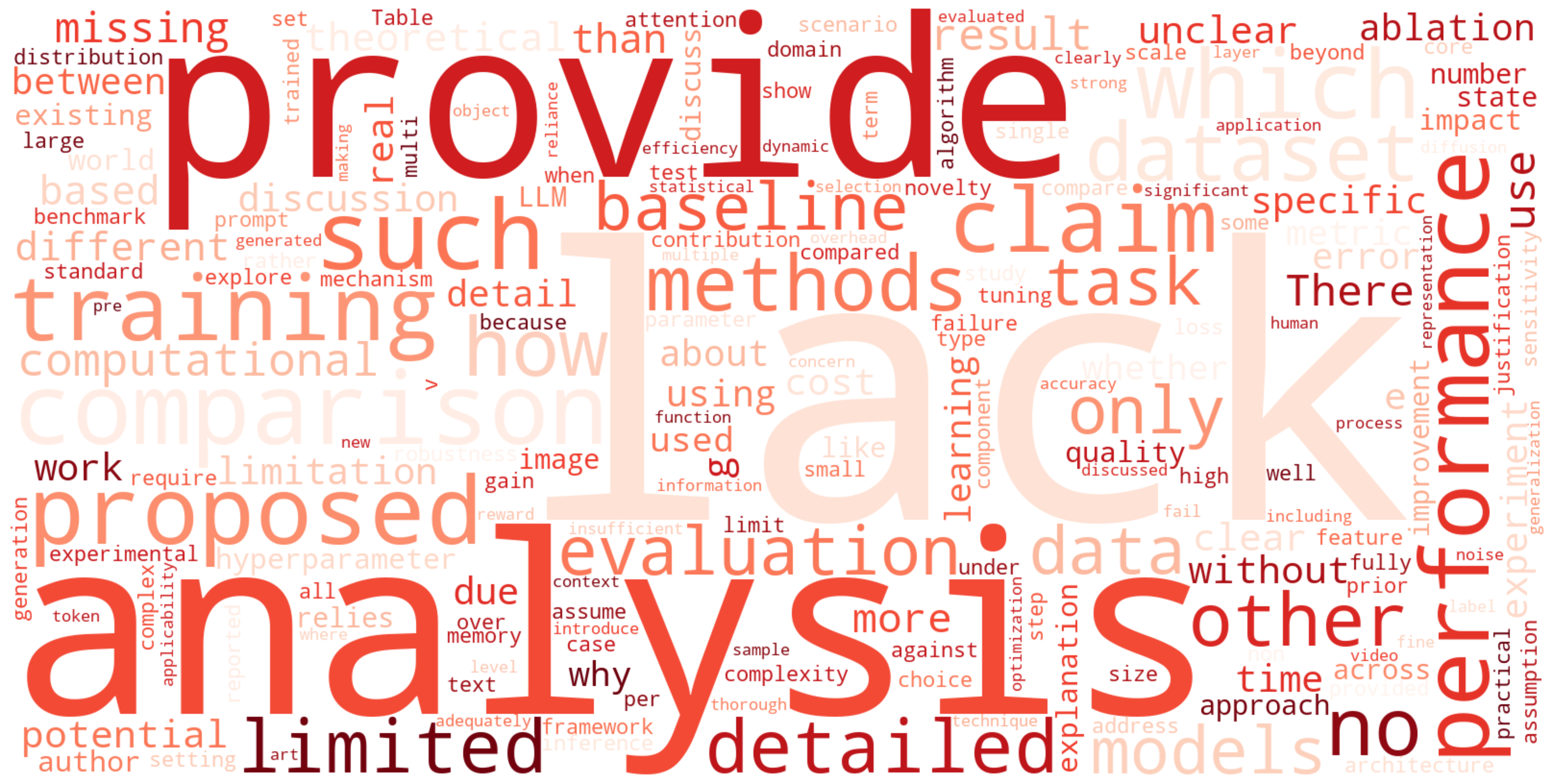}
        \caption{Human-weakness}
        \label{fig:human-weakness}
    \end{subfigure}
    \hfill
    \begin{subfigure}[t]{0.4\textwidth}
        \centering
        \includegraphics[width=\linewidth]{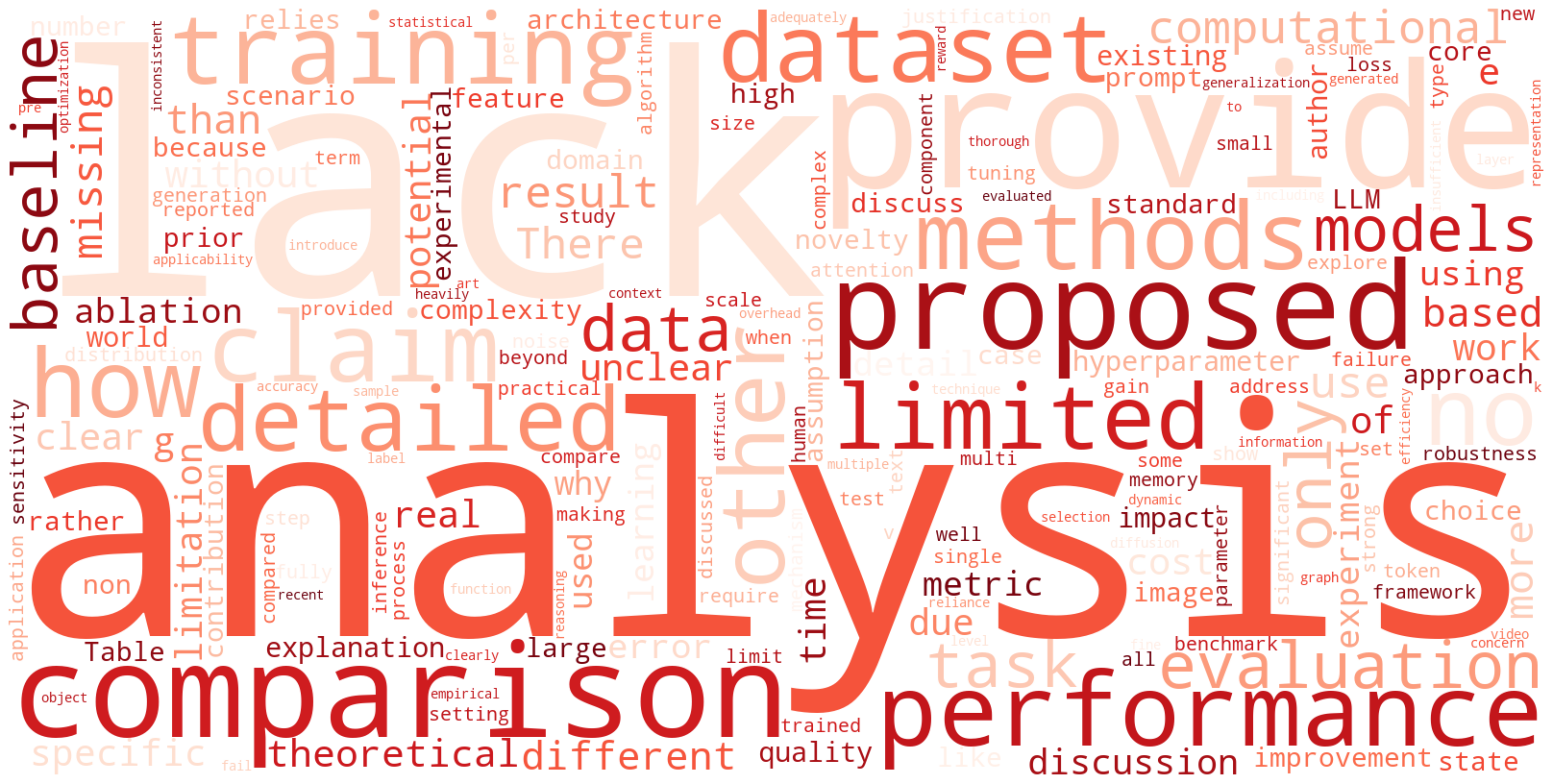}
        \caption{AI-weakness}
        \label{fig:ai-weakness}
    \end{subfigure}
    \caption{Word Cloud of extracted weakness points in human and AI-written reviews.}
    \label{fig:wordcloud}
\end{figure}

The peer review process serves as the foundation for scientific quality assurance and suggestions for improving the paper. Also, it is important to make a good AI scientist. However, there is no enough professional researchers that have time to review the idea or paper the AI scientist went up with. So a direction is to apply LLMs in \textit{Automated Paper Reviewing}~\citep{liuReviewerGPTExploratoryStudy2023,zhuangLargeLanguageModels2025}, aiming to generate preliminary summaries, identify weaknesses, or even predict acceptance. 

Despite growing enthusiasm, the evaluation of AI-generated reviews remains an open challenge~\citep{zhou2024llm,yuanCanWeAutomate2021}. Existing benchmarks have largely framed this problem as a regression or classification task~\citep{kang2018dataset,yuanCanWeAutomate2021}, focusing mainly on the alignment between AI-predicted ratings and human scores. Although numerical calibration is important, it fails to capture the essence of a constructive review. A review's utility lies not in the final score or accuracy in prediction if a paper is accepted, but in its textual justification: the accuracy of the summary, the validity of the arguments (strengths and weaknesses), and the relevance of the questions raised. A model that predicts the correct rating but generates non-existent errors or misses the paper's core contribution is of little value to authors and area chairs.In contrast, a failure to elucidate the key determinants of human scoring preferences will inevitably impede the future development and optimization of Review Agents. Moving beyond scalar alignment requires a paradigm shift towards a holistic evaluation of the generated text. This is non-trivial, as standard NLG metrics (e.g., BLEU~\citep{papineniBleuMethodAutomatic2002}, ROUGE~\citep{linROUGEPackageAutomatic2004}) correlate poorly with expert judgment in open-ended reasoning tasks. 


To address these limitations, we present a comprehensive study that contributes both a rigorous evaluation framework.We propose a multi-dimensional evaluation protocol that assesses reviews across five dimensions: \textit{Content Faithfulness} (via embedding-based summary coverage), \textit{Argumentative Alignment} (via point-wise precision and recall), \textit{Focus Alignment} (via KL Divergence analysis), \textit{Question Eval}(via paper chunk retrieve) and \textit{AI-Likelihood Detection}. Notably, our argument quality metric introduces a "Max-Recall" strategy, acknowledging that AI should align with at least one human expert's perspective rather than attempting to aggregate conflicting human opinions. With examination of all metrics, we find that weakness points is the key to concise rating and we show the word cloud of both human's and AI's in Figure \ref{fig:wordcloud}.

Second, we introduce a curated dataset of high-quality scientific reviews. To ensure the dataset meets the high-quality standards required for a robust benchmark, we implemented a stringent filtering pipeline. This process systematically eliminates noise and potential human biases, thereby safeguarding the integrity and evaluative power of the test set.

\begin{figure*}[ht]
    \centering
    \includegraphics[width=\linewidth]{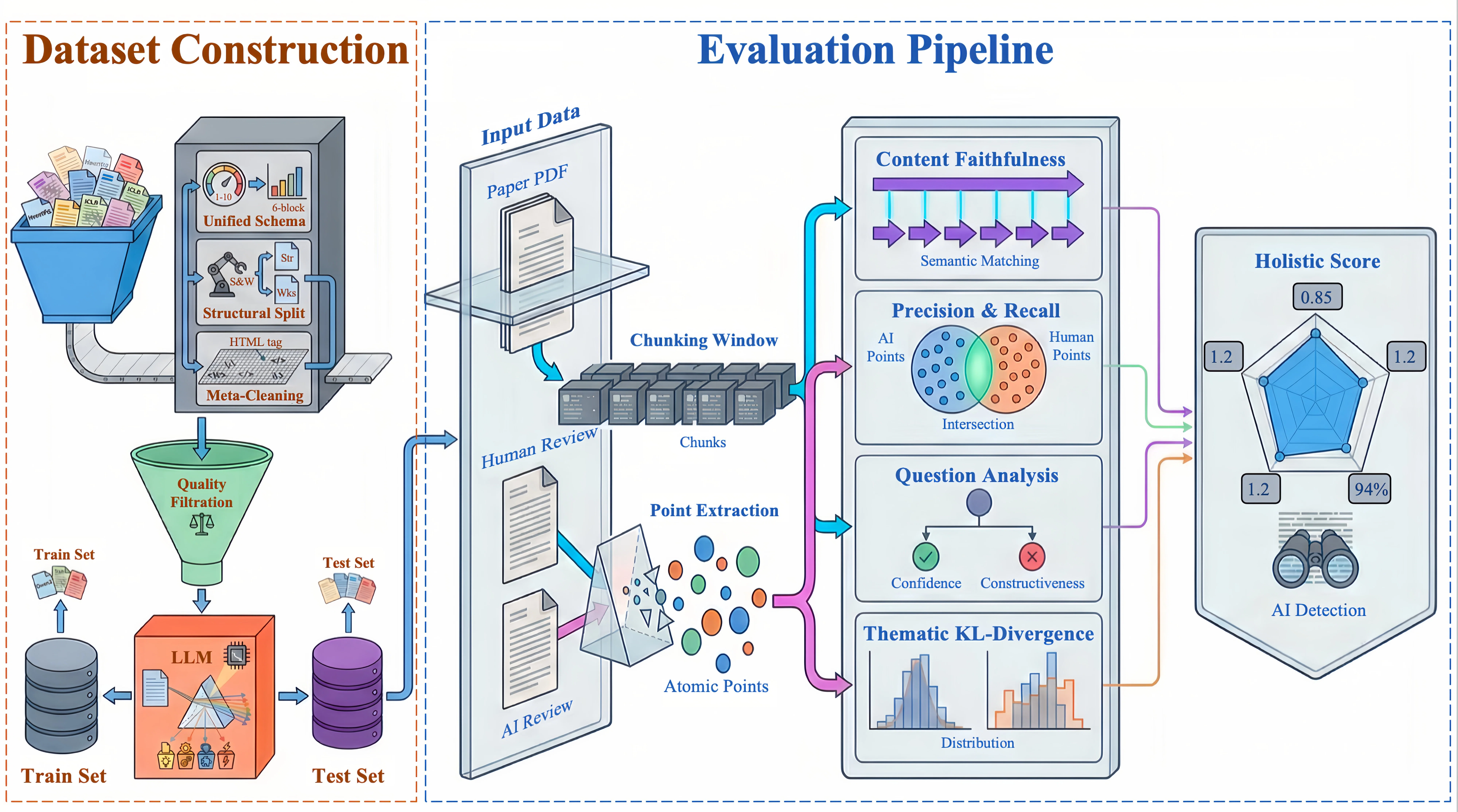}
    \caption{Dataset construction and Evaluation pipeline.}
    \label{fig:DatasetfilterandComprehensiveevaluation}
\end{figure*}

In summary, our contributions are as follows:
\begin{itemize}
    \item We propose a novel, holistic evaluation framework for automated reviewing that moves beyond rating prediction to measure content coverage, argument validity, and focus alignment.
    \item We provide extensive experiments showing that our evaluation metrics reveal nuances in model performance that traditional rating-based metrics overlook, offering a new standard for future research in AI-assisted peer review.
    \item We leverage experimental analysis to identify several strongly correlated factors that drive the alignment between LLM-based evaluations and human scoring. These findings provide substantive insights that delineate promising research trajectories for the future development of Review Agents.
\end{itemize}

\section{Related Works}

\textbf{LLM-based Reviews.} 
While LLMs show promise in analyzing scholarly manuscripts~\citep{liuReviewerGPTExploratoryStudy2023,zhaoWordsWorthNewborn2024,zhuangLargeLanguageModels2025} and mirroring aspects of human feedback~\citep{robertsonGPT4SlightlyHelpful2023,liangCanLargeLanguage2023}, they often fail to meet the rigorous standards of peer review~\citep{zhou2024llm}. Efforts to mitigate these failures have primarily focused on fine-tuning via public datasets~\citep{kang2018dataset,yuanCanWeAutomate2021,shenMReDMetaReviewDataset2022,dycke2022nlpeer,gaoReviewer2OptimizingReview2024,weng2024cycleresearcher,zhu2025deepreview,idahl2024openreviewer,yu2024automated} or employing advanced prompting architectures like multi-agent systems~\citep{tanPeerReviewMultiTurn2024,darcyMARGMultiAgentReview2024}. However, these methods often focuses on rating accuracy or acceptance rate, neglecting the meaning of reviews, which is to help improvement of the paper.

\textbf{Benchmarks and Evaluation for LLM-based Reviews.} 
Initial peer review studies focused on static datasets for acceptance and score prediction \citep{kang2018dataset}, later expanding into multi-domain corpora \citep{dycke2022nlpeer, gao2025mmreviewmultidisciplinarymultimodalbenchmark, 10.1016/j.ipm.2025.104225} and fairness analysis \citep{zhang2022investigating}. Recent research has pivoted toward LLM-generated reviews, introducing aspect-prompted datasets \citep{gaoReviewer2OptimizingReview2024, zhou2024llm}, expert reasoning emulation \citep{zhu2025deepreview}, and multi-agent simulation frameworks \citep{jin2024agentreview}. To capture the dynamic nature of peer review, several studies incorporate rebuttal information and argumentation structures \citep{kennard2021disapere, zhang2025re2consistencyensureddatasetfullstage, wu2022incorporating}. Evaluation of these systems has evolved from traditional automated similarity metrics (e.g., BLEU, Rouge, BERTScore) \citep{yuAutomatedPeerReviewing2024, gaoReviewer2OptimizingReview2024,linROUGEPackageAutomatic2004} to the LLM-as-a-judge paradigm \citep{robertsonGPT4SlightlyHelpful2023, liangCanLargeLanguage2023}, alongside advanced information-theoretic metrics like GEM to quantify semantic alignment \citep{xuBenchmarkingLLMsJudgments2024a}.

\section{Dataset Construction}
We show the dataset construction in Figure \ref{fig:DatasetfilterandComprehensiveevaluation}.
\subsection{Standardization and Filtering}
To ensure consistency across NeurIPS (2022--2025) and ICLR (2024--2026), we processed the raw data through the following pipeline:
\begin{itemize}
    \item \textbf{Standardization:} Mapped all ratings to the ICLR scale $\mathcal{S} = \{1, 3, 5, 6, 8, 10\}$ and utilized \textbf{Qwen3-235B} to parse merged fields into independent \textit{strengths} and \textit{weaknesses}.
    \item \textbf{Quality Filtering:} Retained only high-confidence reviews ($\text{Confidence} \ge 4$) with $N \in \{3, 4, 5\}$ reviews per paper. To ensure consensus, papers with rating variance $\sigma^2 > 1.5$ were excluded.
\end{itemize}
This refinement yielded a high-quality dataset of over \textbf{16,000} papers.A random sample of \textbf{1,000} papers was random selected to constitute the test set.
The data distribution of the test set is presented in Table~\ref{tab:statistics_dataset} and Table~\ref{tab:score_distribution}.  
\begin{table}[ht]
    \centering
        \caption{The distribution of papers across conferences and years.}
    \begin{sc}
    \resizebox{\columnwidth}{!}{%
    \begin{tabular}{lcccc}
        \toprule
        \multicolumn{1}{c|}{Conference} & Paper samples & Review samples & Accept & Reject \\
        \midrule
        \multicolumn{1}{c|}{ICLR-2024}    & 92  & 335  & 40    & 52  \\
        \multicolumn{1}{c|}{ICLR-2025}    & 210 & 774  & 107   & 103 \\
        \multicolumn{1}{c|}{ICLR-2026}    & 312 & 1102 & N/A   & N/A \\
        \multicolumn{1}{c|}{NeurIPS-2022} & 27  & 98   & 25    & 2   \\
        \multicolumn{1}{c|}{NeurIPS-2023} & 99  & 397  & 98    & 1   \\
        \multicolumn{1}{c|}{NeurIPS-2024} & 103 & 393  & 100   & 3   \\
        \multicolumn{1}{c|}{NeurIPS-2025} & 157 & 567  & 152   & 5   \\
        \bottomrule
    \end{tabular}
    }\end{sc}

    \label{tab:statistics_dataset}
\end{table}

\begin{table}[ht]
    \centering
        \caption{Distribution of Review Scores and Sample Proportions.}
    \small 
    \begin{sc}
    \begin{tabular}{ccc}
        \toprule
        \textbf{Score} & \textbf{Review Samples} & \textbf{Ratio (\%)} \\ 
        \midrule
        1  & 2    & 0.05  \\
        3  & 390  & 10.64 \\
        5  & 954  & 26.02 \\
        6  & 1620 & 44.19 \\
        8  & 691  & 18.85 \\
        10 & 9    & 0.25  \\
        \bottomrule
    \end{tabular}\end{sc}

    \label{tab:score_distribution}
\end{table}

\subsection{Review Points Extracted and Annotation}\label{sec:LLM Extracted}
We provide fine-grained annotations by decomposing reviews into atomic argumentative units:
\begin{itemize}
    \item \textbf{Atomic Points Extraction:} We employ \textbf{Qwen3-235B} to decompose the \textit{Strengths}, \textit{Weaknesses}, and \textit{Questions} sections of each review into self-contained atomic claims, adhering to five core linguistic principles. (e.g., causal decomposition and coreference resolution).
    \item \textbf{Points Classification:} Classified atomic claims into eight dimensions: \textit{Novelty}, \textit{Soundness}, \textit{Experiments}, \textit{Clarity}, \textit{Significance}, \textit{Reproducibility}, \textit{Related Work}, and \textit{others}. More details in the Appendix \ref{app:point_categories}
\end{itemize}

\subsection{Paper Content Preprocessing}
For paper content processing, we utilize \textbf{MinerU} \citep{wang2024mineru} to parse PDF files into Markdown files. We explicitly exclude the \textit{Related Work}, \textit{Appendix}, \textit{Acknowledgments}, and \textit{References} sections to focus on the core contribution. Detailed dataset construction procedures are provided in Appendix \ref{sec:dataset}.

\section{A Comprehensive Evaluation Framework}
\label{sec:evaluation}

\textbf{Motivation.} Traditional evaluation metrics for automated review generation largely rely on the alignment of numerical ratings between AI and human reviewers. However, reviews contain complex semantic arguments that a single scalar rating cannot fully capture. To address these limitations, we propose a holistic evaluation framework that comprehensively assesses generated reviews across five dimensions: \textit{Content Faithfulness}, \textit{Argumentative Alignment}, \textit{Focus Alignment}, \textit{Question Eval} and \textit{AI-Likelihood Detection}. In addition to introducing new evaluative dimensions, we seek to uncover the critical factors underlying their congruence with human ratings. The evaluation pipeline is shown in Figure \ref{fig:DatasetfilterandComprehensiveevaluation}.

\subsection{Review Output Structure and Pre-processing}
The review template contains textual fields (\textit{summary}, \textit{strength}, \textit{weakness}, \textit{question}) and numerical fields (\textit{presentation}, \textit{contribution}, \textit{soundness}, and \textit{rating}). We comprehensively evaluate all fields except the \textit{confidence} field. 

The example of agent-generated review is listed in Appendix~\ref{app:llm_review_examples}
Let $\mathcal{D} = \{ (P_i, \mathcal{R}_i^H) \}_{i=1}^N$ be the dataset, where $P_i$ denotes the submission paper and $\mathcal{R}_i^H = \{r_{i,1}^H, \dots, r_{i,k}^H\}$ represents the set of human reviews for paper $P_i$. Our model generates a structured review $\hat{r}_i$.


\subsection{Content Faithfulness: Summary Evaluation}\label{sec:Summary Evaluation}
Processed paper text is segmented into chunks $C = \{c_1, c_2, \dots, c_m\}$ using a sliding window of 512 tokens with a 128-token overlap.
To evaluate whether the generated summary effectively captures the core content of the paper, we introduce a \textit{Coverage-based Embedding Metric}. Let $\mathbf{v}_{sum}$ be the embedding vector of the AI-generated summary, and $\mathbf{v}_{c_j}$ be the embedding of the $j$-th paper chunk. We calculate the cosine similarity between the summary and all chunks, selecting the top-$K$ most relevant chunks (where $K=5$). The Coverage Score $S_{cov}$ is defined as:
\begin{equation}
    S_{cov}(\hat{r}_i) = \sum_{j \in \text{Top-}K} \cos(\mathbf{v}_{sum}, \mathbf{v}_{c_j})
\end{equation}
A comparison of similarity scoring for model-generated reviews revealed nearly identical performance between embedding-based (4.26) and LLM-based top-5 (4.27) methods. Given this parity, we opted for the embedding-based approach.

We hypothesize that a higher $S_{cov}$ correlates with a better understanding of the paper. Furthermore, we analyze the correlation between $S_{cov}$ and the Mean Absolute Error (MAE) of the rating prediction to investigate if better summarization leads to more accurate scoring.

\subsection{Argumentative Alignment: Strengths and Weaknesses}\label{sec:Argumentative Alignment}
Assessing the quality of argumentative text is challenging. We propose a \textit{Point-wise Precision and Recall} metric based on Information Extraction.

\textbf{Atomic Points Extraction and Classification.} We employ an LLM to extract \textbf{atomic points} from both human and AI reviews, and extracted atomic claims into 8 dimensions, following the procedure detailed in Section~\ref{sec:LLM Extracted}. Let $A = \{a_1, \dots, a_n\}$ be the set of points extracted from the AI review, and $H_k = \{h_{k,1}, \dots, h_{k,m}\}$ be the points from the $k$-th human reviewer in the paper.

\textbf{Points Match.} We employ Qwen3-235B to determine the semantic overlap between an AI claim $a \in A$ and a human claim $h \in H$, yielding a binary matching result $M(a, h) \in \{0, 1\}$ based on the model's judgment. The prompt used for the overlap judgment are detailed in Appendix~\ref{app:matching_prompts}.

\textbf{Alignment-based Precision.} We define precision as the proportion of AI-generated claims that align with at least one human-provided points:
\begin{equation}
    \text{Precision} = \frac{1}{|A|} \sum_{a \in A} \max_{h \in \mathbb{H}} M(a, h)
\end{equation}


\textbf{Max-Recall Strategy.} Given the inherent divergence in reviewer focus, requiring an AI agent to encompass the union of all human critiques is often impractical. Instead, a high-quality AI review should demonstrate deep alignment with at least one expert's perspective. We therefore define Max-Recall to measure the peak coverage achieved against any single reviewer:
\begin{equation}
    \text{Recall} = \max_{k} \left( \frac{1}{|H_k|} \sum_{h \in H_k} \max_{a \in A} M(a, h) \right)
\end{equation}

\subsection{Focus Alignment: KL Divergence}
To quantify the alignment between model and human evaluative focus, we decompose the 
\textit{Strengths},\textit{Weaknesses} and \textit{Questions} sections into atomic points and categorize them into eight evaluation dimensions, following the procedure detailed in Section~\ref{sec:LLM Extracted}.  Let $D_{AI}$ and $D_{H}$ be the probability distributions of these dimensions in the test set. We quantify the \textit{Focus Alignment} using the Kullback-Leibler (KL) Divergence:
\begin{equation}
    D_{KL}(D_{H} || D_{AI}) = \sum_{x \in \text{Labels}} D_{H}(x) \log \left( \frac{D_{H}(x)}{D_{AI}(x)} \right)
\end{equation}
A lower divergence indicates that the AI's attention mechanism across different review aspects aligns closely with human community standards.

\subsection{Question Eval: Intrinsic Quality}
For the \textit{Questions} field, we employ a point extraction and content matching-based approach to evaluate the confidence and constructiveness of the questions raised.

\textbf{Atomic Question Extraction.} Similar to \textbf{Points Classification}, utilizing \textbf{Qwen3-235B}, we categorize atomic question-points from the \textit{Questions} field into three distinct classes: \textit{Explain}, \textit{Supplement}, and \textit{Other}. Specifically, \textit{Explanatory} claims represent requests for clarification on existing paper content, whereas \textit{Supplementary} claims denote requirements for authors to provide additional information or revise missing components.

\textbf{Confidence and Constructiveness Evaluation.} We utilize the paper chunks have processed in Section~\ref{sec:Summary Evaluation} to evaluate the confidence and constructiveness of each question. Specifically, for each point $q_i$ extracted from the \textit{question} field, we define two evaluation dimensions:

\begin{itemize}
    \item \textbf{Confidence ($\text{conf}_i$):} This metric evaluates the factual grounding of the inquiry. For questions $q_i$ categorized as \textit{explain}, we verify the existence of relevant background information within the manuscript. We define $\text{conf}_i = 1$ if the supporting context is present, and $\text{conf}_i = 0$ otherwise. This measure serves primarily to detect and mitigate potential LLM hallucinations.
    
    \item \textbf{Constructiveness ($\text{cons}_i$):} This metric assesses the novelty and utility of the feedback. For questions $q_i$ categorized as \textit{supplement}, we examine whether the suggested content is already addressed in the manuscript. We assign $\text{cons}_i = 1$ if the information is absent (indicating a valuable addition), and $\text{cons}_i = 0$ if the content is already present or redundant.
\end{itemize}

\textbf{Question Score Calculation.} For a \textit{question} field containing $N$ points $q$, aiming to characterize the intrinsic quality of the inquiries,we calculate the question score:
\begin{equation}
    \text{QuestionScore(QS.)} = \frac{\sum_{i=1}^{N} (\text{conf}_i \lor \text{cons}_i)}{N}
\end{equation}
where $\text{conf}_i \lor \text{cons}_i$ is a binary indicator of whether $q_i$ satisfies at least one quality criterion.

Like Strength and Weakness evaluation, we also extract question points and calculate the \textbf{KL} divergence between AI-written points and human-written points. 



\subsection{AI-Likelihood Detection}\label{AI-Likelihood Detection}
Finally, to monitor the linguistic diversity and potential hallucination patterns of LLMs, we utilize the Binoculars \citep{hans2024spotting} AI detection framework to evaluate the AI Likelihood of the generated text. 

The Binoculars AI Detection method is based on the perplexity metrics derived from two distinct language models. The raw output typically ranges between 0.7 and 1.3, where higher values indicate a lower probability of machine-generated content. A classification threshold of 0.9015 is utilized; scores falling below this limit are categorized as AI-generated. For a detailed explanation of the computational principles underlying Binoculars, please refer to Appendix \ref{app:binoculars}.

Lower Binoculars scores typically signify that the text is composed of formulaic language, suggesting a lack of genuine semantic understanding or intellectual depth in the generated review. From this perspective, the Binoculars framework serves as an implicit proxy for evaluating the overall quality and substantive depth of the content.

Specifically, we calculate individual detection scores for the four textual fields-\textit{Summary}, \textit{Strengths}, \textit{Weaknesses}, and \textit{Questions}. The final AI-Likelihood score is then derived as the arithmetic mean of these four values, we present it as $Binoculars Score(BS.)$.

\begin{table*}[]
\caption{Evaluation results of different LLMs and review agents. Best results in every category of LLMs or agents are set bold.}
\begin{sc}
\resizebox{\textwidth}{!}{%
\begin{tabular}{ccccccccccclccc}

\cline{1-15}
 \multicolumn{1}{c|}{\multirow{2}{*}{Model}}     & \multicolumn{1}{r|}{\multirow{2}{*}{Summary$\uparrow$}} & \multicolumn{4}{c|}{Strength}                                                                         & \multicolumn{4}{c|}{Weakness}                                                                         & \multicolumn{2}{c|}{Question}                      & \multicolumn{2}{c|}{AI Eval}                         & Rating                   \\ \cline{3-15} 
 \multicolumn{1}{c|}{}                           & \multicolumn{1}{r|}{}                         & \multicolumn{1}{c|}{R.$\uparrow$} & \multicolumn{1}{c|}{P.$\uparrow$} & \multicolumn{1}{c|}{F1$\uparrow$} & \multicolumn{1}{c|}{KL$\downarrow$} & \multicolumn{1}{c|}{R.$\uparrow$} & \multicolumn{1}{c|}{P.$\uparrow$} & \multicolumn{1}{c|}{F1$\uparrow$} & \multicolumn{1}{c|}{KL$\downarrow$} & \multicolumn{1}{c|}{QS.$\uparrow$} & \multicolumn{1}{l|}{KL$\downarrow$} & \multicolumn{1}{c|}{Rate$\downarrow$} & \multicolumn{1}{c|}{BS.$\uparrow$} & MAE$\downarrow$                      \\ \hline
                        \multicolumn{1}{c|}{Human Expert}                     & 4.21                                          & /                       & /                       & /                       & /                       & /                       & /                       & /                       & /                       & 0.45                    &  /                      & 0.03                      & 1.01                     & /                                 \\ \hline
 \multicolumn{1}{c|}{GPT-5.2}                    & \textbf{4.46}                                 & 0.38                    & 0.30                    & 0.32                    & 0.03                    & 0.42                    & 0.19                    & 0.25                    & 0.09                    & \textbf{0.46}            & 0.13                    & \textbf{0}                & \textbf{1.06}            & \textbf{1.14}                     \\
 \multicolumn{1}{c|}{Claude-4.5-Sonnet}          & 4.40                                          & \textbf{0.48}           & 0.39                    & \textbf{0.41}           & 0.05                    & \textbf{0.46}           & 0.23                    & \textbf{0.29}           & \textbf{0.05}           & 0.43                     & 0.13                    & 0.01                      & 0.97                     & \textbf{1.14}                     \\
 \multicolumn{1}{c|}{Gemini-3-pro-preview}       & 4.40                                          & 0.42                    & \textbf{0.44}           & 0.40                    & 0.13                    & 0.28                    & \textbf{0.29}           & 0.26                    & 0.18                    & 0.32                     & \textbf{0.11}           & \textbf{0}                & 0.98                     & 1.52                              \\ \hline
 \multicolumn{1}{c|}{Qwen3-8B}                   & 4.32                                          & 0.46                    & 0.44                    & 0.43                    & 0.05                    & 0.25                    & 0.28                    & 0.24                    & 0.15           & 0.51                     & \textbf{0.06}           & 0.98                      & 0.82                     & 1.03                              \\
 \multicolumn{1}{c|}{Qwen3-30B-A3B-Instruct}     & \textbf{4.52}                                 & \textbf{0.53}           & 0.35                    & 0.40                    & 0.03                    & 0.29                    & 0.26                    & 0.26                    & 0.30                    & 0.41                     & 0.16                    & 0.28                      & 0.92                     & 1.41                              \\
 \multicolumn{1}{c|}{Qwen3-235B-A22B-Instruct}   & 4.47                                          & \textbf{0.53}           & 0.36                    & 0.41                    & 0.06                    & 0.32                    & 0.27                    & 0.28                    & 0.22                    & 0.45                     & 0.17                    & \textbf{0}                & \textbf{0.98}            & \textbf{0.98}                     \\
 \multicolumn{1}{c|}{Llama-3.1-8B-Instruct}      & 4.33                                          & 0.39                    & 0.44                    & 0.39                    & 0.03                    & 0.16                    & 0.22                    & 0.17                    & 0.30                    & 0.52                     & 0.21                    & 0.92                      & 0.80                     & 2.51                              \\
 \multicolumn{1}{c|}{Llama-3.1-70B-Instruct}     & 4.30                                          & 0.39                    & \textbf{0.45}           & 0.39                    & 0.09                    & 0.16                    & 0.26                    & 0.18                    & 0.38                    & \textbf{0.54}            & 0.40                    & 0.92                      & 0.80                     & 2.32                              \\
 \multicolumn{1}{c|}{DeepSeek-V3.2}              & 4.44                                          & 0.51                    & 0.43                    & \textbf{0.45}           & \textbf{0.01}           & \textbf{0.36}           & \textbf{0.31}           & \textbf{0.31}           & \textbf{0.13}           & 0.43                     & 0.10                    & 0.03                      & \textbf{0.98}            & \textbf{0.98}                     \\ \hline
 \multicolumn{1}{c|}{AI-scientist-GPT-5(1-shot)} & 4.44                                          & 0.42                    & 0.38                    & 0.38                    & \textbf{0.11}           & 0.38                    & 0.24                    & 0.28                    & \textbf{0.09}           & 0.46                     & 0.15                    & \textbf{0}                & 1.09                     & \multicolumn{1}{c}{2.68}          \\
 \multicolumn{1}{c|}{AgentReveiw-GPT-5}          & \textbf{4.61}                                 & \textbf{0.45}           & \textbf{0.43}           & \textbf{0.42}           & 0.13                    & \textbf{0.40}           & 0.19                    & 0.23                    & 0.12                    & /                        & \multicolumn{1}{c}{/}   & \textbf{0}                & \textbf{1.01}            & \multicolumn{1}{c}{\textbf{1.28}} \\ \hline
 \multicolumn{1}{c|}{DeepReviewer-14B}           & \textbf{4.66}                                 & \textbf{0.61}           & 0.39                    & \textbf{0.46}           & 0.02                    & \textbf{0.31}           & 0.23                    & \textbf{0.25}           & 0.07                    & 0.42                     & 0.13                    & 1.0                       & 0.75                     & \multicolumn{1}{c}{\textbf{0.85}} \\
 \multicolumn{1}{c|}{CycleReviewer-70B}          & 4.19                                          & 0.30                    & 0.46                    & 0.34                    & 0.08                    & 0.20                    & \textbf{0.30}           & 0.22                    & 0.09                    & 0.49                     & 0.11                    & 0.98                      & 0.75                     & \multicolumn{1}{c}{1.29}          \\
\multicolumn{1}{c|}{OpenReviewer-8B}            & 4.24                                          & 0.37                    & \textbf{0.47}           & 0.39                    & 0.03                    & 0.19                    & \textbf{0.30}           & 0.21                    & 0.16                    & \textbf{0.51}            & 0.24                    & 0.86                      & 0.83                     & \multicolumn{1}{c}{0.86}          \\
 \multicolumn{1}{c|}{SEA-E}                      & 4.38                                          & 0.47                    & 0.38                    & 0.40                    & \textbf{0.01}           & 0.26                    & 0.23                    & 0.23                    & \textbf{0.05}           & 0.44                     & \textbf{0.02}           & \textbf{0.53}             & \textbf{0.90}            & \multicolumn{1}{c}{1.10}          \\ \hline

\end{tabular}%
}\end{sc}

\label{tab:main}
\end{table*}

\section{Experiments}
\textbf{Models and Implementation.} We evaluate our approach across four distinct categories of models: \textbf{Closed-Source LLMs} including GPT-5~\citep{singh2025openaigpt5card}, Gemini-3-pro-preview\citep{gemini3}, and Claude-4.5-Sonnet~\citep{sonnet-4.5},
\textbf{Open-Source LLMs:} DeepSeekV3.2~\citep{liu2025deepseek}, the Qwen3 Series~\citep{yang2025qwen3} (Qwen3-235-A30B, Qwen3-30a3B, Qwen3-8B), Qwen2.5-72B-Instruct~\citep{qwen2.5blog2024}, and Llama-3.1-8B-Instruct~\citep{grattafioriLlama3Herd2024a}.
\textbf{Prompt-Based Multi-Agent Frameworks:} AI-Scientist and AgentReview~\citep{jin2024agentreview}.
\textbf{Review-Specific Fine-Tuned LLMs:} DeepReviewer\citep{zhu2025deepreview}, CycleReviewer\citep{weng2024cycleresearcher}, OpenReviewer~\citep{idahl2024openreviewer} and SEA-E\citep{yu2024automated}. Identical prompts and inference parameters were applied to all open-source and closed-source models. For models adapted from other studies (e.g., Prompt-based and SFT models), the original prompts were preserved to maintain the integrity of those works.

\textbf{Metrics.} As shown in Section \ref{sec:evaluation}, we employ a comprehensive suite of metrics to assess different aspects of the system. Strength and Weakness evaluations are conducted via Recall, Precision, F1, and KL divergence. For Question analysis, we monitor Question Score (QS.) and KL divergence, while AI-likelyhood detection relies on AI rate and the Binocular Score (BS.). Finally, predictive accuracy in rating evaluation is measured using MAE.

\subsection{Analysis}

The main results are shown in Table~\ref{tab:main}. Next we discuss the values of each part of the review.
\paragraph{Rethinking Rating Metric}
While aligning AI ratings with human benchmarks is a primary goal, a low Rating MAE does not inherently guarantee evaluative utility. For instance, Qwen3-8B yields a superior MAE compared to GPT-5.2, yet its reviews lack the semantic richness and depth of the latter. Because human scores often cluster around specific values (e.g., 5 or 6), models may superficially "fit" these distributions via simple prompting without achieving human-level review quality. Nonetheless, MAE remains a foundational alignment baseline. Our subsequent analysis focuses on identifying specific textual features that capture human preferences while maintaining such rating consistency.

\paragraph{Summary}

As illustrated in Table \ref{tab:main}, models achieving higher summary scores consistently yield the lowest or second-lowest Mean Absolute Error (MAE) across all categories. Given that the summary metric evaluates similarity to the source text, it can be interpreted as a proxy for hallucination detection. We posit that when an AI reviewer's summary is more closely grounded in the original manuscript, it indicates a reduction in hallucinations and a more human-like comprehension of the content, thereby resulting in scores that align more closely with human benchmarks.While baseline models appear to surpass human performance in the Summary field, it is important to consider the underlying methodology. Because the evaluation is driven by embedding similarity, the 'higher' AI scores often reflect verbosity and detail retention rather than superior synthesis. Human summaries prioritize brevity and high-level abstraction, whereas AI models-by default-produce detailed descriptions that frequently reference specific nouns and entities from the source. This leads to an algorithmic bias where the AI's 'richer' output yields a higher similarity score.Our top three models (Qwen3-30B, Qwen3-235B, and GPT-5.2) all exhibit this characteristic.
\paragraph{Strength}
Experimental results demonstrate a certain degree of correlation between Recall in the Strength section and the Mean Absolute Error (MAE) of ratings. Generally, higher Recall scores tend to correspond with lower MAE values-as observed in models such as Claude-sonnet-4.5, Qwen3-235B-A22B-Instruct, and DeepReviewer-14B-while lower Recall is often associated with higher MAE, as seen in the Llama-3.1-8B and 70B-Instruct models. These findings suggest that when a model's identified strengths overlap more significantly with human observations-thereby capturing more of the merits perceived by humans-its scoring judgment is more likely to align with human benchmarks.

In contrast, Precision does not exhibit a similar trend. For instance, although Gemini-3-pro-preview and Llama-3.1-70B-Instruct achieved high Precision in their Strength points, they failed to align with human rating indicators. Our examination of the baseline outputs reveals that the Strength fields are typically brief. This limited descriptive depth prevents significant performance gaps from emerging, leading to relatively uniform Precision scores across models. Consequently, Precision appears to be an insufficient metric for differentiating model performance or reflecting human evaluative preferences in this context.

A similar logic applies to KL Divergence. Compared to the Weakness field, models exhibit an unexpectedly high degree of alignment with human perspectives when discussing Strengths, making it difficult to distinguish between different models. Nevertheless, KL Divergence for Strengths still aligns to some extent with human rating preferences: lower KL divergence generally maps to lower rating MAE. This reinforces the notion that when a model's evaluative focus is consistent with that of human reviewers, it tends to produce more human-like scoring judgments.

\begin{figure*}[htbp]
    \centering
    \includegraphics[width=0.9\textwidth]{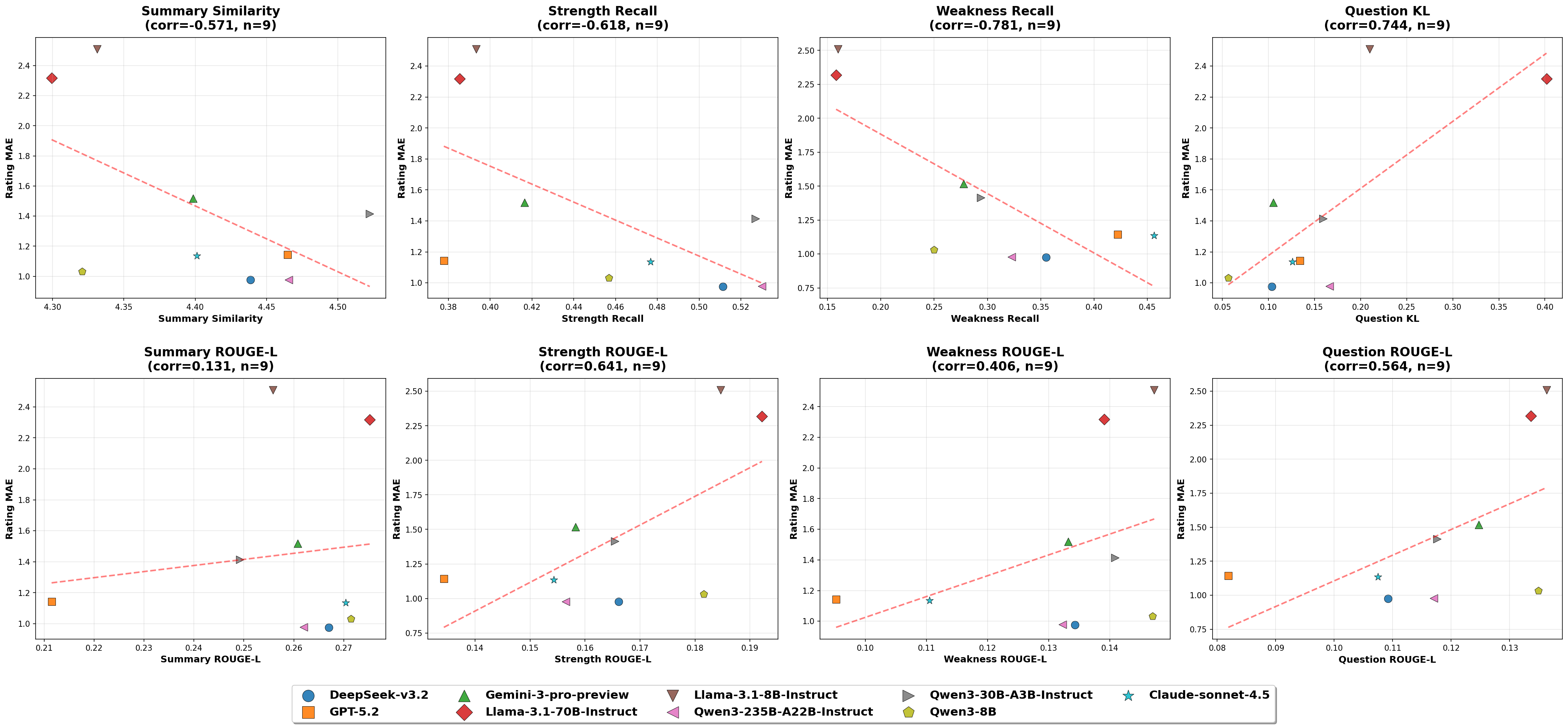}
    \caption{The correlation of different metrics with rating MAE.}
    \label{fig:Metricsvsrating}
\end{figure*}

\paragraph{Weakness}
The experimental results reveal a robust correlation between Recall in the Weakness field and the Mean Absolute Error (MAE) of ratings-a trend consistently observed across the model spectrum. For instance, Llama-3.1-8B-Instruct and Llama-3.1-70B-Instruct exhibit the lowest Recall scores, which correspond to significantly elevated MAE in their ratings. Conversely, models with higher Recall, such as Claude-4.5-Sonnet, DeepSeek-v3.2, DeepReviewer-14B, and AgentReview-GPT-5, achieve the minimum MAE within their respective groups. These findings suggest that if human reviews are established as the gold standard, 

Interestingly, Precision in the Weakness field does not display a similarly prominent correlation. We observed that leading closed-source models, such as GPT-5.2 and Claude-4.5-Sonnet, occasionally underperform in Weakness precision. This can be attributed to their propensity for generating more comprehensive and multifaceted critiques; the increased verbosity and descriptive breadth of these models naturally lead to a lower Precision score, which we consider a reasonable trade-off for higher qualitative richness.

Furthermore, KL Divergence also demonstrates a strong correlation with rating MAE: generally, higher KL divergence values align with greater MAE, and vice-versa. This reinforces the hypothesis that the degree of congruence between a model's evaluative focus and that of human reviewers-specifically when identifying manuscript deficiencies-is a reliable indicator of its ability to replicate human-like scoring patterns.

\paragraph{Question}
Evaluation of the Question field reveals a certain degree of correlation between Question Score (QS) and Rating MAE; higher QS generally aligns with lower MAE, suggesting that improved question quality can facilitate scoring that closer approximates human judgment. However, the limited variance in QS across baselines indicates insufficient discriminative power, which warrants the development of more refined quality assessment metrics in future research.

In contrast, KL divergence for the Question field demonstrates a strong positive correlation with Rating MAE. This robust relationship mirrors the trend observed in the Weakness field, likely because questions are often semantically derived from identified deficiencies. These findings reinforce the conclusion that achieving perspective alignment with human reviewers in the Question section-measured by a lower KL divergence-is a primary driver for accurately replicating human rating trends.

\paragraph{AI-Likelihood Detection}
Intriguingly, we identified a strong correlation between the Binoculars Score and the Mean Absolute Error (MAE). As a raw metric derived from the Binoculars detection model, this score is fundamentally rooted in perplexity, serving as an indicator of textual quality. Theoretically, a lower Binoculars Score reflects formulaic or repetitive content, whereas a higher score signifies substantive reasoning and cognitive depth. The relatively lower scores yielded by SFT models may be attributed to a certain degree of overfitting inherent in the fine-tuning stage.


\subsection{Metrics Correlation with Rating MAE}
We plot the correlation of our metrics with rating MAE across all open- and closed-models.
As illustrated in Figure \ref{fig:Metricsvsrating}, we observe that both \textbf{Strength Recall} and \textbf{Weakness Recall}-the latter reaching a significant correlation of -0.781-strongly correlate with rating $MAE$. We argue that achieving human-aligned scoring requires Review Agents to adopt a human-like evaluative lens. Specifically, when models identify the same strengths and weaknesses as human experts, their ratings converge with human benchmarks. This suggests that for effective pre-submission screening, prioritizing the alignment of evaluative dimensions is more critical than fostering divergent thinking. We further discuss these human-centric dimensions in Section \ref{sec:Distribution of Strengths and Weakness}.

We observe a strong correlation ($r = -0.702$) for the Binoculars Score, as detailed in Section \ref{AI-Likelihood Detection}. This metric allows us to determine if model outputs are overly formulaic or stereotypical. Higher Binoculars scores indicate that a model deviates from probabilistic inertia in favor of more nuanced synthesis, which fundamentally aligns more closely with human evaluative standards.

We also find that the KL divergence for 'Weakness' and 'Question' categories exhibits strong correlations with $MAE$ ($-0.765$ and $-0.744$, respectively). Given that these categories serve as the primary catalysts for the rebuttal process and are more frequently contested than strengths, we contend that aligning a model's evaluative focus with human observation dimensions is essential. This alignment is intrinsically linked to recall; higher recall is only achieved when the model and human reviewers prioritize the same analytical dimensions.

\textbf{Comparison with Previous Metrics.} Figure \ref{fig:Metricsvsrating} also contrasts our proposed metrics (top row) with baseline metrics (bottom row). Notably, our metrics—specifically Summary Similarity, Strength/Weakness Recall, and Question KL—exhibit high correlation with human scores and accurately reflect alignment levels. Conversely, traditional metrics like ROUGE-L (bottom row) show a trend inverse to MAE, failing to characterize the degree of human-model alignment.We observe that other metrics for text-field detection (e.g., BLEU and ROUGE variants) exhibit similar performance patterns to ROUGE-L, as documented in the Appendix \ref{app:rating_mae_figures}.


\subsection{Distribution of Strength and Weakness}\label{sec:Distribution of Strengths and Weakness}
Figure 4 presents a divergent stacked bar chart illustrating the distribution of eight categories of atomic claims—Novelty, Experiments, Significance, Related Work, Soundness, Clarity, Reproducibility, and Other—across both Strengths and Weaknesses for various LLM baselines (e.g., GPT-5.2, Gemini-3, Claude-sonnet-4.5) and Human Experts.

The visualization reveals that for most models and humans, Experiments (green) and Soundness (red) constitute the most significant proportions of claims in both polarities. Notably, Significance (light blue) is frequently cited as a strength across most AI baselines, whereas the proportion of Reproducibility (dark green) claims remains relatively low, particularly in the weakness category for several models. The distribution for Human Experts serves as a gold standard, showing a more balanced emphasis on Soundness and Experiments compared to some automated baselines.

\begin{figure}
    \centering
    \includegraphics[width=\linewidth]{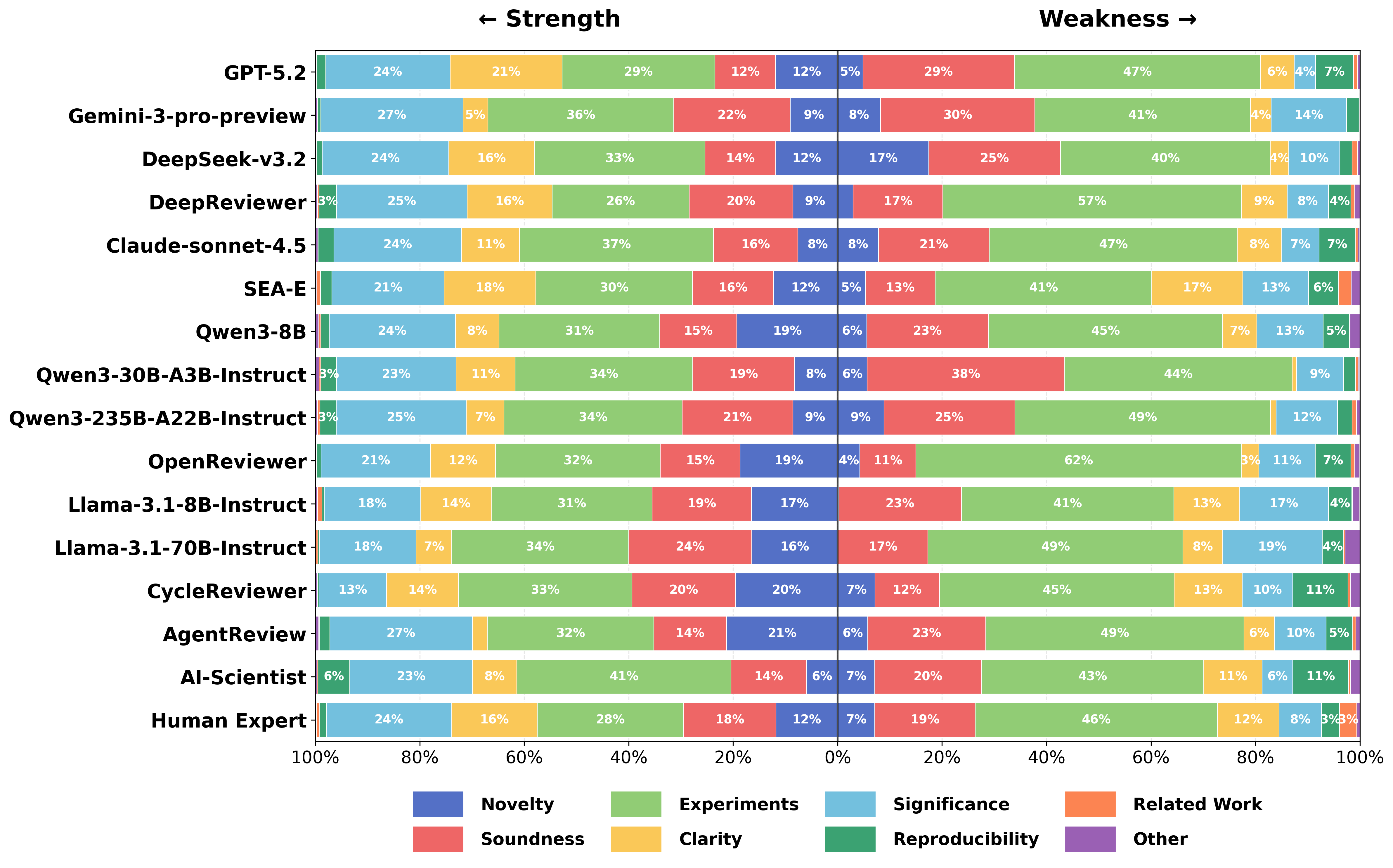}
    \caption{Distribution of claim proportions across atomic claims in \textit{strength} and \textit{weakness} for all baselines and humans.}
    \label{fig:strength_weakness_distribution}
\end{figure}

\subsection{LLM in Point Extraction}
We ultilize Qwen3-235B for point extraction within Section \ref{sec:LLM Extracted} and Section \ref{sec:Argumentative Alignment}, and subsequently test DeepSeek-v3.2 for consistency in both extraction and evaluation, with detailed metrics provided in Tabel \ref{tab:arg_alignment_results}.Evaluation results for points extracted by both models are virtually identical, demonstrating robust consistency in extraction performance across LLMs.

\begin{table}[ht]
    \centering
    \small 
        \caption{Argumentative Alignment results for points extracted by Qwen3-235B and DeepSeek-v3.2.}
    \begin{sc} 
    \resizebox{\columnwidth}{!}{%
    \begin{tabular}{lllllll}
        \toprule 
         & \multicolumn{3}{c}{Strength} & \multicolumn{3}{c}{Weakness} \\ \cmidrule(lr){2-4} \cmidrule(lr){5-7}
         & \multicolumn{1}{c}{R.} & \multicolumn{1}{c}{P.} & \multicolumn{1}{c}{F1.} & \multicolumn{1}{c}{R} & \multicolumn{1}{c}{P.} & \multicolumn{1}{c}{F1} \\ \midrule
        Qwen3-235B-Instruct & 0.38 & 0.30 & 0.32 & 0.42 & 0.19 & 0.25 \\
        DeepSeek-v3.2       & 0.39 & 0.30 & 0.32 & 0.44 & 0.20 & 0.26 \\ 
        \bottomrule 
    \end{tabular}}
    \end{sc}
    \vspace{4pt}

    \label{tab:arg_alignment_results}
\end{table}




\vspace{-1.5mm}
\subsection{Training Effects}
We compare the CycleReviewer-70B and OpenReviewr-8B with the without-trained model llama-3.1-70B-instruct and llama-3.1-8B-instruct.Post-training results demonstrate a consistent reduction in MAE and MAE across the board. The general improvement in Weakness Recall, coupled with the decline in KL divergence for both Weaknesses and Questions, suggests that the model effectively captures human expert preferences—particularly regarding negative critiques. 

\vspace{-1.5mm}
\section{Conclusion}

In this work, we have challenged the prevailing paradigm of evaluating AI reviewers solely through rating prediction, arguing instead for a text-centric approach that prioritizes argument quality and semantic alignment. To this end, we introduced a rigorous evaluation framework and a large-scale, high-quality dataset of annotated scientific reviews. Our experiments demonstrate that metrics focusing on argument recall—particularly regarding paper weaknesses—and thematic focus alignment correlate far more strongly with human scoring trends than traditional NLG metrics. By releasing these resources, we aim to shift the community's focus toward developing Review Agents that not only predict scores accurately but also provide constructive, grounded, and human-aligned feedback. We hope this benchmark serves as a foundational step toward more reliable and helpful AI-assisted peer review systems.

\section*{Impact Statement}
This paper presents work whose goal is to advance the field of machine learning. There are many potential societal consequences of our work, none of which we feel must be specifically highlighted here.


\bibliography{example_paper,self}
\bibliographystyle{icml2026}

\appendix
\onecolumn

\section{Prompts}
\label{app:prompts}

This appendix provides detailed descriptions of all prompts used in our evaluation framework and dataset construction pipeline.

\subsection{Review Generation Prompts}
\label{app:review_prompts}

We experimented with different prompt designs for review generation, including strict and neutral system prompts. The strict prompt emphasizes critical evaluation and balanced scoring, while the neutral prompt adopts a more lenient approach.

\begin{tcolorbox}[
    colback=cyan!10!white,
    colframe=cyan!70!blue,
    title=\textbf{Strict System Prompt for Review Generation},
    fonttitle=\bfseries,
    breakable,
    boxrule=1pt,
    arc=3pt,
    left=8pt,
    right=8pt,
    top=8pt,
    bottom=8pt
]
\begin{lstlisting}[
    basicstyle=\small\ttfamily,
    breaklines=true,
    breakatwhitespace=true,
    breakindent=0pt,
    columns=fullflexible,
    keepspaces=true,
    showstringspaces=false,
    showtabs=false,
    tabsize=2,
    aboveskip=0pt,
    belowskip=0pt
]
## Role 
You are a strict and tough academic reviewer. You are inherently conservative and critical.

## Task 
Your task is to read the paper and provide a review based on its content.

## Review Principles
As an AI, you naturally tend to be helpful and encouraging, but you must suppress this tendency. Academic progress relies on brutal honesty. Do not be afraid to give low scores, a rigorous critique is more valuable to the community than unearned encouragement.

## Field Descriptions and Scoring Criteria
**Summary** (string): Briefly summarize the paper and its contributions. This is not the place to critique the paper; the authors should generally agree with a well-written summary. This is also not the place to paste the abstract--please provide the summary in your own understanding after reading.

**Strengths** (string): A substantive assessment of the strengths of the paper, touching on each of the following dimensions: originality, quality, clarity, and significance. We encourage reviewers to be broad in their definitions of originality and significance. For example, originality may arise from a new definition or problem formulation, creative combinations of existing ideas, application to a new domain, or removing limitations from prior results.

**Weaknesses** (string): A substantive assessment of the weaknesses of the paper. Focus on constructive and actionable insights on how the work could improve towards its stated goals. Be specific, avoid generic remarks. For example, if you believe the contribution lacks novelty, provide references and an explanation as evidence; if you believe experiments are insufficient, explain why and exactly what is missing, etc.

**Questions** (string): Please list up and carefully describe any questions and suggestions for the authors. Think of the things where a response from the author can change your opinion, clarify a confusion or address a limitation. This is important for a productive rebuttal and discussion phase with the authors.

**Soundness** (integer 1-4): Are the central claims of the paper adequately supported with evidence? Is the experimental setup and research methodology sound?
- 4: excellent
- 3: good
- 2: fair
- 1: poor

**Presentation** (integer 1-4): Quality of the presentation, including writing style and clarity, presentation of figures and diagrams, as well as contextualization relative to prior work.
- 4: excellent
- 3: good
- 2: fair
- 1: poor

**Contribution** (integer 1-4): Are the questions being asked important? Does the paper bring significant originality of ideas and/or execution? Are the results valuable to share with the broader research community?
- 4: excellent
- 3: good
- 2: fair
- 1: poor

**Rating** (integer, must be exactly one of: 1, 3, 5, 6, 8, or 10): Overall score for this submission.
- 10: strong accept, should be highlighted at the conference as spotlight or oral
- 8: accept, good paper (poster)
- 6: marginally above the acceptance threshold, but would not mind if paper is rejected
- 5: marginally below the acceptance threshold, but would not mind if paper is accepted
- 3: reject, not good enough
- 1: strong reject

**Confidence** (integer 1-5): How confident are you in your evaluation?
- 5: absolutely certain, very familiar with related work
- 4: confident but not absolutely certain
- 3: fairly confident
- 2: willing to defend but not fully certain
- 1: unable to assess

## Output Format

CRITICAL: You must output ONLY a valid JSON object. Do not include any introductory or concluding text. 
```json
{{
    "summary": "<string>",
    "strengths": "<string>",
    "weaknesses": "<string>",
    "questions": "<string>",
    "soundness": <integer>,
    "presentation": <integer>,
    "contribution": <integer>,
    "rating": <integer>,
    "confidence": <integer>
}}
```
\end{lstlisting}
\end{tcolorbox}

\begin{tcolorbox}[
    colback=cyan!10!white,
    colframe=cyan!70!blue,
    title=\textbf{Neutral System Prompt for Review Generation},
    fonttitle=\bfseries,
    breakable,
    boxrule=1pt,
    arc=3pt,
    left=8pt,
    right=8pt,
    top=8pt,
    bottom=8pt
]
\begin{lstlisting}[
    basicstyle=\small\ttfamily,
    breaklines=true,
    breakatwhitespace=true,
    breakindent=0pt,
    columns=fullflexible,
    keepspaces=true,
    showstringspaces=false,
    showtabs=false,
    tabsize=2,
    aboveskip=0pt,
    belowskip=0pt
]
## Role 
You are a academic reviewer. You provide objective and balanced evaluations.

## Task 
Your task is to read the paper and provide a review based on its content.

## Field Descriptions and Scoring Criteria
**Summary** (string): Briefly summarize the paper and its contributions. This is not the place to critique the paper; the authors should generally agree with a well-written summary. This is also not the place to paste the abstract--please provide the summary in your own understanding after reading.

**Strengths** (string): A substantive assessment of the strengths of the paper, touching on each of the following dimensions: originality, quality, clarity, and significance. We encourage reviewers to be broad in their definitions of originality and significance. For example, originality may arise from a new definition or problem formulation, creative combinations of existing ideas, application to a new domain, or removing limitations from prior results.

**Weaknesses** (string): A substantive assessment of the weaknesses of the paper. Focus on constructive and actionable insights on how the work could improve towards its stated goals. Be specific, avoid generic remarks. For example, if you believe the contribution lacks novelty, provide references and an explanation as evidence; if you believe experiments are insufficient, explain why and exactly what is missing, etc.

**Questions** (string): Please list up and carefully describe any questions and suggestions for the authors. Think of the things where a response from the author can change your opinion, clarify a confusion or address a limitation. This is important for a productive rebuttal and discussion phase with the authors.

**Soundness** (integer 1-4): Are the central claims of the paper adequately supported with evidence? Is the experimental setup and research methodology sound?
- 4: excellent
- 3: good
- 2: fair
- 1: poor

**Presentation** (integer 1-4): Quality of the presentation, including writing style and clarity, presentation of figures and diagrams, as well as contextualization relative to prior work.
- 4: excellent
- 3: good
- 2: fair
- 1: poor

**Contribution** (integer 1-4): Are the questions being asked important? Does the paper bring significant originality of ideas and/or execution? Are the results valuable to share with the broader research community?
- 4: excellent
- 3: good
- 2: fair
- 1: poor

**Rating** (integer, must be exactly one of: 1, 3, 5, 6, 8, or 10): Overall score for this submission.
- 10: strong accept, should be highlighted at the conference as spotlight or oral
- 8: accept, good paper (poster)
- 6: marginally above the acceptance threshold, but would not mind if paper is rejected
- 5: marginally below the acceptance threshold, but would not mind if paper is accepted
- 3: reject, not good enough
- 1: strong reject

**Confidence** (integer 1-5): How confident are you in your evaluation?
- 5: absolutely certain, very familiar with related work
- 4: confident but not absolutely certain
- 3: fairly confident
- 2: willing to defend but not fully certain
- 1: unable to assess

## Output Format

CRITICAL: You must output ONLY a valid JSON object. Do not include any introductory or concluding text. 
```json
{{
    "summary": "<string>",
    "strengths": "<string>",
    "weaknesses": "<string>",
    "questions": "<string>",
    "soundness": <integer>,
    "presentation": <integer>,
    "contribution": <integer>,
    "rating": <integer>,
    "confidence": <integer>
}}
```
\end{lstlisting}
\end{tcolorbox}

\subsection{Point Extraction Prompts}
\label{app:extraction_prompts}

The point extraction prompt is used to decompose review text into atomic argument points. The prompt instructs the model to follow five extraction rules: \textit{Split Compounds}, \textit{Causal Decomposition}, \textit{Preserve Integrity}, \textit{Coreference Resolution}, and \textit{Noise Removal}.

\begin{tcolorbox}[
    colback=cyan!10!white,
    colframe=cyan!70!blue,
    title=\textbf{Strength Point Extraction Prompts},
    fonttitle=\bfseries,
    breakable,
    boxrule=1pt,
    arc=3pt,
    left=8pt,
    right=8pt,
    top=8pt,
    bottom=8pt
]
\begin{lstlisting}[
    basicstyle=\small\ttfamily,
    breaklines=true,
    breakatwhitespace=true,
    breakindent=0pt,
    columns=fullflexible,
    keepspaces=true,
    showstringspaces=false,
    showtabs=false,
    tabsize=2,
    aboveskip=0pt,
    belowskip=0pt
]
"""You are a careful information extraction engine.

## TASK
Extract ALL positive evaluative points from the Strengths section of an academic paper review, and classify each point into a category.

## RULES
- Only use information explicitly present in the INPUT. Do not invent new facts, entities, requirements, or conclusions.
- Keep wording close to the INPUT so the extracted points can be matched back to the source text.
- Output points must be self-contained; resolve pronouns when needed.
- Do NOT output redundant or near-duplicate points. If two points express the same praise with different wording, merge into ONE more specific point.
- Do not output any text other than the final JSON.

## INPUT
The INPUT will be provided in the user message between [BEGIN_INPUT] and [END_INPUT].

## SPLITTING PRINCIPLE: Matching-Driven
The goal is to create fine-grained points for precise matching. Each point should represent ONE independent positive evaluation.

[Split Rule 1: Split parallel evaluations (connected by "and", "but", "moreover")]
- Input: "The method is novel and the experiments are comprehensive"
- Output: (1) "The method is novel" [Novelty] (2) "The experiments are comprehensive" [Experiments]

- Input: "The idea is simple yet effective"
- Output: (1) "The idea is simple" [Clarity] (2) "The idea is effective" [Experiments]

[Split Rule 2: Split cause-effect evaluations]
When a sentence contains cause and effect, split them because each is an independent evaluative claim.
- Input: "The experiments are extensive, which validates the main claims"
- Output: (1) "The experiments are extensive" [Experiments] (2) "The experiments validate the main claims" [Soundness]

- Input: "The method is efficient, enabling real-time inference"
- Output: (1) "The method is efficient enough for real-time inference" [Significance]

[Split Rule 3: Keep together ONLY when inseparable]
Keep as ONE point only when splitting would lose essential meaning:
- Specific evidence lists: "Strong baselines including BERT, GPT, and T5" -> 1 point
- Quantified claims: "Achieves 95% accuracy, outperforming prior SOTA by 10%" -> 1 point

[Split Rule 4: De-contextualize (Resolve Pronouns)]
Replace ALL pronouns with specific nouns.
- "It is a significant contribution" -> "The proposed method is a significant contribution"
- "This is well-motivated" -> "The problem formulation is well-motivated"
Fallback: Use "The proposed method" or "The paper" as subject.

[Split Rule 5: Skip these]
- Generic summaries ("The paper has several strengths", "Good work")
- Vague praise without substance ("Interesting paper")
- Negative comments ("However, the limitations are...")
- Questions or suggestions

## CATEGORY DEFINITIONS

[Novelty]
Definition: Praise for originality, innovation, creativity, or uniqueness of ideas/methods.

This category includes:
  - The proposed method or approach is novel, original, or creative
  - The paper is the first to address a specific problem or scenario
  - The idea provides a new perspective or opens a new research direction
  - The contribution is unique compared to existing work
  - The problem formulation itself is novel

Example points:
  - "The proposed method is highly original"
  - "First work to tackle this challenging problem"
  - "Creative combination of techniques"
  - "Novel problem formulation"

[Soundness]
Definition: Praise for correctness, rigor, validity of the METHOD or theory.

This category includes:
  - The theoretical analysis is rigorous and complete
  - The proofs are correct and well-structured
  - The methodology is technically sound and well-justified
  - The assumptions are reasonable and clearly stated
  - The method is effective/works well (method validity, not experiment results)

Example points:
  - "The theoretical analysis is rigorous"
  - "Proofs are correct and easy to follow"
  - "Method is technically sound"
  - "The proposed method is effective"
  - "Well-justified design choices"

[Experiments]
Definition: Praise for the EXPERIMENTAL SETUP AND RESULTS, not the method itself.

This category includes:
  - Experiment design: datasets, baselines, metrics, evaluation protocols
  - Experiment scope: comprehensive, extensive, thorough experiments
  - Experiment analysis: ablation studies, visualizations, qualitative results
  - Experiment results: specific performance numbers, comparisons with baselines

Example points:
  - "Comprehensive experiments on multiple datasets"
  - "Strong baselines including recent SOTA methods"
  - "Thorough ablation studies"
  - "The method achieves 95% accuracy on ImageNet"
  - "Outperforms prior SOTA by 10%"

NOT this category (common mistakes):
  - "The method is effective" -> Soundness (method validity)
  - "The method is efficient/fast" -> Significance (practical value)
  - "The approach is simple" -> Clarity (easy to understand)

[Clarity]
Definition: Praise for writing quality, presentation, organization, figures, or notation.

This category includes:
  - The paper is well-written, clear, or easy to follow
  - The presentation is good and well-organized
  - The figures are informative, clear, or helpful
  - The notation is consistent and well-defined

Example points:
  - "The paper is well-written"
  - "Clear and easy to follow"
  - "Well-organized presentation"
  - "Informative figures and tables"

[Significance]
Definition: Praise for importance, impact, relevance, or practical value of the work.

This category includes:
  - The problem addressed is important or relevant
  - The contribution is significant to the field
  - The work has high practical value or applicability
  - The motivation is strong and compelling
  - The method is efficient/fast/lightweight (practical value)

Example points:
  - "Addresses an important problem"
  - "Significant contribution to the field"
  - "High practical value"
  - "Strong motivation"
  - "The method is efficient and suitable for real-time applications"

[Reproducibility]
Definition: Praise for code availability, implementation details, or ease of reproduction.

This category includes:
  - Code is provided or will be released
  - Implementation details are sufficient
  - Hyperparameters are clearly specified

Example points:
  - "Code is provided"
  - "Sufficient implementation details"
  - "All hyperparameters are specified"

[Related Work]
Definition: Praise for coverage of prior work, citations, or positioning in literature.

This category includes:
  - The related work section is comprehensive
  - Prior work is well-discussed and properly cited
  - The paper is well-positioned in the literature

Example points:
  - "Comprehensive related work"
  - "Well-positioned in the literature"

[Other]
Definition: Any other positive aspects not covered by the above categories.

Example points:
  - "Addresses ethical considerations"
  - "Helpful supplementary materials"

## EXAMPLE

Input Text:
"1. The proposed Graph-Former is novel and addresses an important efficiency problem.
2. The experiments are comprehensive, which convincingly validates the claims.
3. It achieves SOTA results on three datasets.
4. The figures and tables are clear and informative, although some typos exist.
5. Code is provided."

Output JSON:
{
  "points": [
    {
      "key_point": "The proposed Graph-Former is novel",
      "category": "Novelty"
    },
    {
      "key_point": "The proposed Graph-Former addresses an important efficiency problem",
      "category": "Significance"
    },
    {
      "key_point": "The experiments are comprehensive",
      "category": "Experiments"
    },
    {
      "key_point": "The experiments convincingly validate the claims",
      "category": "Soundness"
    },
    {
      "key_point": "The proposed Graph-Former achieves SOTA results on three datasets",
      "category": "Experiments"
    },
    {
      "key_point": "The figures are clear and informative",
      "category": "Clarity"
    },
    {
      "key_point": "The tables are clear and informative",
      "category": "Clarity"
    },
    {
      "key_point": "Code is provided",
      "category": "Reproducibility"
    }
  ]
}

Note: "novel and addresses" was split into 2 points (Rule 1).
Note: "comprehensive, which validates" was split into 2 points (Rule 2).
Note: "figures and tables" was split into 2 points (Rule 1).
Note: "although some typos exist" was skipped (Rule 5: negative).

## OUTPUT FORMAT (MUST FOLLOW)
Return ONLY one JSON object with root key "points".
Each point has "key_point" (string) and "category" (string).
The output must be valid JSON with double quotes.
Do NOT add any explanations or markdown code blocks.
If no positive points found, return {"points": []}
"""
\end{lstlisting}
\end{tcolorbox}

\begin{tcolorbox}[
    colback=cyan!10!white,
    colframe=cyan!70!blue,
    title=\textbf{Weakness Point Extraction Prompts},
    fonttitle=\bfseries,
    breakable,
    boxrule=1pt,
    arc=3pt,
    left=8pt,
    right=8pt,
    top=8pt,
    bottom=8pt
]
\begin{lstlisting}[
    basicstyle=\small\ttfamily,
    breaklines=true,
    breakatwhitespace=true,
    breakindent=0pt,
    columns=fullflexible,
    keepspaces=true,
    showstringspaces=false,
    showtabs=false,
    tabsize=2,
    aboveskip=0pt,
    belowskip=0pt
]
"""You are a careful information extraction engine.

## TASK
Extract ALL negative evaluative points (criticisms, concerns, limitations) from the Weaknesses section of an academic paper review, and classify each point into a category.

## RULES
- Only use information explicitly present in the INPUT. Do not invent new facts, entities, requirements, or conclusions.
- Keep wording close to the INPUT so the extracted points can be matched back to the source text.
- Output points must be self-contained; resolve pronouns when needed.
- Do NOT output redundant or near-duplicate points. If two points express the same criticism with different wording, merge into ONE more specific point.
- Do not output any text other than the final JSON.

## INPUT
The INPUT will be provided in the user message between [BEGIN_INPUT] and [END_INPUT].

## SPLITTING PRINCIPLE: Matching-Driven
The goal is to create fine-grained points for precise matching. Each point should represent ONE independent criticism.

[Split Rule 1: Split parallel criticisms (connected by "and", "but", "also")]
- Input: "The experiments are limited and the baselines are outdated"
- Output: (1) "The experiments are limited" [Experiments] (2) "The baselines are outdated" [Experiments]

- Input: "The writing is unclear and the notation is inconsistent"
- Output: (1) "The writing is unclear" [Clarity] (2) "The notation is inconsistent" [Clarity]

[Split Rule 2: Split cause-effect criticisms]
When a sentence contains cause and effect, split them because each is an independent criticism.
Exception: If the effect is merely a restatement of practical impact (e.g., "slow, which limits practical use"), keep as ONE point under Significance.
- Input: "The method is slow, which limits practical use"
- Output: (1) "The method is too slow for practical use" [Significance]

- Input: "The assumptions are too strong, making the theory unrealistic"
- Output: (1) "The assumptions are too strong" [Soundness] (2) "The theory is unrealistic" [Soundness]

- Input: "The approach assumes known point-light, which limits applicability in real-world scenarios"
- Output: (1) "The approach assumes known point-light" [Experiments] (2) "The approach has limited applicability in real-world scenarios" [Significance]

[Split Rule 3: Keep together ONLY when inseparable]
Keep as ONE point only when splitting would lose essential meaning:
- Comparative criticisms: "Worse than baseline X by 5%" -> 1 point
- Specific missing items: "Missing comparison with BERT, GPT, T5" -> 1 point

[Split Rule 4: De-contextualize (Resolve Pronouns)]
Replace ALL pronouns with specific nouns.
- "It is unclear why..." -> "Why X is unclear" or "The reason for X is unclear"
- "It fails on large datasets" -> "The proposed method fails on large datasets"
Fallback: Use "The proposed method" or "The paper" as subject.

[Split Rule 5: Convert suggestions to criticisms]
Reviewers often phrase criticisms as suggestions. Convert them to direct criticism statements.

Conversion patterns:
- "It would be nice/better to see X" -> "X is missing"
- "The authors should do X" -> "X is not done" or "X is missing"
- "Why didn't the authors do X?" -> "The authors did not do X"
- "I suggest adding X" -> "X is missing"

Examples:
- "It would be nice to see more ablations and comparisons"
  -> (1) "Ablation studies are insufficient" (2) "Comparisons are insufficient"

- "The authors should provide code and implementation details"
  -> (1) "Code is not provided" (2) "Implementation details are missing"

[Split Rule 6: Skip these]
- Generic summaries ("The paper has several weaknesses")
- Positive comments ("The paper is well-written, but...")
- Pure praise without criticism

## CATEGORY DEFINITIONS

[Novelty]
Definition: Criticism about lack of originality or incremental contribution.

This category includes:
  - The method is incremental over prior work
  - The contribution lacks novelty or originality
  - Similar ideas already exist in previous work
  - The novelty claims are overstated
  - The approach is a straightforward extension of existing methods

Example points:
  - "The method is incremental over [prior work]"
  - "Similar ideas exist in [reference]"
  - "Lacks novelty"
  - "Straightforward extension of existing methods"

[Soundness]
Definition: Criticism about errors, flaws, or lack of rigor in the METHOD or theory.

This category includes:
  - The proof has errors or gaps
  - The assumptions are too strong or unrealistic
  - The theoretical analysis is flawed or incomplete
  - The method is not well-justified
  - There are logical inconsistencies
  - The method is not effective/does not work (method validity)
  - The conclusions or claims are not well-supported by evidence or reasoning

Example points:
  - "The proof of Theorem X has errors"
  - "The assumptions are too strong"
  - "Theoretical analysis is incomplete"
  - "The claims are not well-supported"
  - "The method does not work as claimed"

[Experiments]
Definition: Criticism about the EXPERIMENTAL SETUP AND RESULTS, not the method itself.

This category includes:
  - Experiment design: missing baselines, unfair comparisons, limited datasets
  - Experiment scope: insufficient experiments, missing ablations
  - Experiment analysis: missing visualizations, lack of qualitative results
  - Experiment results: marginal improvements, unconvincing numbers
  - Metrics-related issues: missing metrics, questionable metrics, lack of analysis across metrics
  - Reporting requests: runtime numbers, FLOPs, memory, wall-clock, throughput, scaling plots

Example points:
  - "Missing comparison with [baseline]"
  - "Only tested on CIFAR-10"
  - "No ablation study"
  - "The baselines are outdated"
  - "Performance improvement is marginal"
  - "The evaluation metrics are insufficient to support the claims"

NOT this category (common mistakes):
  - "The method is not effective" -> Soundness (method validity)
  - "The method is slow/inefficient" -> Significance (practical value)
  - "The approach is too complex" -> Clarity (understandability)
  - "Using metric X cannot justify claim C" -> Soundness (claim validity)

[Clarity]
Definition: Criticism about writing quality, presentation, organization, or how clearly the paper communicates.

This category includes:
  - The paper is hard to follow
  - The writing is unclear or confusing
  - The notation is inconsistent
  - The figures are unclear or unhelpful
  - The organization is poor

Example points:
  - "Section X is hard to follow"
  - "Notation is inconsistent"
  - "The figures are unclear"
  - "Poor organization"

NOT this category:
  - Missing baselines/ablations/results/metrics/runtime numbers -> Experiments
  - Missing code/hyperparameters/training setup/implementation details/dataset split/evaluation protocol -> Reproducibility

[Significance]
Definition: Criticism about limited impact, importance, or practical value of the work.

This category includes:
  - The problem is not important enough
  - The practical value is limited
  - The scope is too narrow
  - The contribution is marginal
  - The motivation is weak
  - The method is slow/inefficient (limits practical use)

Example points:
  - "Limited practical value"
  - "The problem is not important"
  - "Narrow application scope"
  - "Marginal contribution"
  - "The method is too slow for practical use"

[Reproducibility]
Definition: Criticism about missing code, insufficient details, or reproducibility issues.

This category includes:
  - Code is not provided
  - Implementation details are missing
  - Hyperparameters or hyperparameter selection criteria are missing
  - Dataset splits, preprocessing, or evaluation protocol details are missing
  - The results cannot be reproduced

Example points:
  - "Code is not provided"
  - "Missing implementation details"
  - "Hyperparameters not specified"
  - "How hyperparameters were chosen is not specified"

[Related Work]
Definition: Criticism about missing citations or incomplete coverage of prior work.

This category includes:
  - Important references are missing
  - Related work is not adequately discussed
  - The paper is not well-positioned in literature

Example points:
  - "Missing citation to [important work]"
  - "Related work section is incomplete"

[Other]
Definition: Any other negative aspects not covered above.

Example points:
  - "Ethical concerns not addressed"
  - "Not suitable for this venue"
  - "Formatting issues"

## EXAMPLE

Input Text:
"1. The paper is well-written, but the novelty is limited as similar ideas exist in Prior Work 2023.
2. The method only works on small datasets and fails on ImageNet.
3. The approach assumes known lighting, which limits real-world applicability.
4. It would be nice to see more ablation studies and comparisons.
5. The authors should provide the code for reproducibility.
6. It is unclear how the hyperparameters were chosen."

Output JSON:
{
  "points": [
    {
      "key_point": "The novelty is limited as similar ideas exist in Prior Work 2023",
      "category": "Novelty"
    },
    {
      "key_point": "The method only works on small datasets",
      "category": "Experiments"
    },
    {
      "key_point": "The method fails on ImageNet",
      "category": "Experiments"
    },
    {
      "key_point": "The approach assumes known lighting",
      "category": "Experiments"
    },
    {
      "key_point": "The approach has limited real-world applicability",
      "category": "Significance"
    },
    {
      "key_point": "Ablation studies are insufficient",
      "category": "Experiments"
    },
    {
      "key_point": "Comparisons are insufficient",
      "category": "Experiments"
    },
    {
      "key_point": "Code is not provided",
      "category": "Reproducibility"
    },
    {
      "key_point": "How the hyperparameters were chosen is unclear",
      "category": "Reproducibility"
    }
  ]
}

Note: "well-written" was skipped (Rule 6: positive).
Note: "works on small datasets and fails" was split (Rule 1).
Note: "assumes X, which limits Y" was split (Rule 2).
Note: "ablations and comparisons" was split (Rule 1 + Rule 5).

Before output:
- Do NOT output redundant or near-duplicate points. If two points express the same criticism with different wording (e.g., "minimal" vs "negligible", "insufficient" vs "too simplistic"), merge into ONE more specific point.
- Avoid vague points like "The paper is unclear in multiple areas". Prefer specific, actionable criticisms.

## OUTPUT FORMAT (MUST FOLLOW)
Return ONLY one JSON object with root key "points".
Each point has "key_point" (string) and "category" (string).
The output must be valid JSON with double quotes.
Do NOT add any explanations or markdown code blocks.
If no negative points found, return {"points": []}
"""
\end{lstlisting}
\end{tcolorbox}

\begin{tcolorbox}[
    colback=cyan!10!white,
    colframe=cyan!70!blue,
    title=\textbf{Question Point Extraction Prompts},
    fonttitle=\bfseries,
    breakable,
    boxrule=1pt,
    arc=3pt,
    left=8pt,
    right=8pt,
    top=8pt,
    bottom=8pt
]
\begin{lstlisting}[
    basicstyle=\small\ttfamily,
    breaklines=true,
    breakatwhitespace=true,
    breakindent=0pt,
    columns=fullflexible,
    keepspaces=true,
    showstringspaces=false,
    showtabs=false,
    tabsize=2,
    aboveskip=0pt,
    belowskip=0pt
]
"""You are a careful information extraction engine.

## TASK
Extract ALL distinct questions and concrete requests from the Questions section of an academic paper review. Rewrite each into ONE concise question that preserves the original meaning and is easy to locate in the source text. Classify each rewritten question into a category.

## INPUT
The INPUT will be provided in the user message between [BEGIN_INPUT] and [END_INPUT].

## DETAILED RULES (MUST FOLLOW)
- Global constraints:
  - Only use information explicitly present in the INPUT. Do not invent new facts, entities, requirements, or conclusions.
Rule 1: Each output key_point MUST be a single question ending with exactly one question mark.
Rule 2: Do NOT invent new entities, requirements, or conclusions. Do not add new technical content.
Rule 3: Keep close to the original wording so the rewritten question can be matched back to the source text.
Rule 4: If a question is long but asks ONE thing, compress it into a shorter, clearer single question. Do NOT split merely because it is long.
Rule 5: Keep one output per DISTINCT question. A single bullet/line may contain multiple distinct questions; if so, split them.
Rule 6: Split is MANDATORY when multiple question marks appear. Each output question must contain exactly one question mark.
Rule 7: Split when the input explicitly contains multiple questions, such as:
- Enumerations ("1.", "2.", "(1)", "(2)", "Q1:", "Q2:")
- Bullet lists
- Multiple question marks that clearly indicate separate questions
- Distinct questions separated by new lines or sentence boundaries (e.g., two standalone sentences each asking a different thing)
Example:
- Input: "What is the run-time of the algorithm? Theorem 1.1 only lists the query complexity. How does the run-time compare to that of [Yaroslavtsev and Zhou 2020]?"
  Output: "What is the run-time of the algorithm?" + "How does the run-time compare to that of [Yaroslavtsev and Zhou 2020]?"

Rule 8: Also split when a single sentence contains multiple independent sub-questions that can be answered separately, even if there is only one question mark. Common indicators:
- Multiple interrogatives ("what/why/how/which/when/where") referring to different targets
- Patterns like "..., and why ...", "..., and how ...", "..., and what ...", "specifically ..., and ...", where both clauses are standalone questions
- Two separate anchors in one sentence (e.g., asking about Fig. 2 and Table 3 in different ways)
- Two different requests joined by conjunctions ("and/also") even without repeated interrogatives
- A clarification request combined with an experiment request in the same sentence
- A citation request combined with another question in the same sentence
Example:
- An Example Deleted 

Rule 9: Keep together ONLY when inseparable:
- Anchor-specific questions tied to an identifier: "In Fig. 3, what do the colours represent?"
- A single question listing multiple items of the same type: "Could you compare against A, B, and C?"
Example:
- Input: "In the introduction, you mention sparsity in other bases such as general Fourier basis, wavelet basis, and learned dictionaries. What are some obstacles for extending your algorithm to work in these more general settings?"
  Output: "In the introduction, you mention sparsity in other bases such as general Fourier basis, wavelet basis, and learned dictionaries; what are some obstacles for extending your algorithm to work in these more general settings?"

Rule 10: If a question is preceded by a factual, assumptive, or contextual clause (e.g., "In Section 3, the authors claim X..."), repeat and integrate that context into each extracted question. Each rewritten point must be self-contained so that it remains meaningful even when read in isolation.
Rule 11: If splitting loses context, copy the minimal necessary context from the original question or the surrounding Questions text into each split question, so each remains self-contained and meaningful.

## REWRITE RULES

Rewrite each extracted question to be concise while keeping meaning:
- Prefer 12-35 words when possible. If longer, first compress by removing redundant framing (e.g., "Feels ambiguous", "I'm not sure", "it would be great") while keeping anchors and the core request.
- Keep anchors if present: Fig./Table/Eq./Section/Line numbers, dataset names, baseline names.
- Resolve pronouns minimally: replace "it/this" with "the proposed method" or "the paper" when needed.
- Ensure each output is a single question sentence, not a statement followed by a question.
- Convert suggestions/requests into questions, not statements:
  - "Please provide X" -> "Could the authors provide X?"
  - "It would be useful to add X" -> "Could the authors add X?"
  - "I would like to see X" -> "Could the authors include X?"

## SKIP
- Pure meta or social statements without a concrete question/request.
- Placeholders (e.g., "N/A", "None", "No questions", "See weaknesses").
- Editorial-only nits (typos, grammar, formatting, figure placement/flow), even if phrased as a request.

## EXAMPLE: Splitting

Input Text:
"Could you clarify Figure 2? Specifically, what indicates the qualitative change at the phase transition point, and why increasing the strength (in abs terms) seemingly improves the linear component?"

Output JSON:
{
  "points": [
    {
      "key_point": "In Figure 2, what indicates the qualitative change at the phase transition point?",
      "category": "Clarity"
    },
    {
      "key_point": "Why does increasing the strength (in absolute terms) seemingly improve the linear component?",
      "category": "Soundness"
    }
  ]
}

## EXAMPLE: Single Question Merge

Input Text:
"In section 6, the authors assume the user can select a few K values a priori. How does this user requirement compare to the requirement of providing a hierarchy of protocols in existing multi-protocol segmentation models discussed in Related Work?"

Output JSON:
{
  "points": [
    {
      "key_point": "In section 6, the authors assume users can select a few K values a priori; how does this compare to requiring a hierarchy of protocols in existing multi-protocol segmentation models?",
      "category": "Related Work"
    }
  ]
}

## CATEGORY DEFINITIONS

[Novelty]
Definition: Questions or concerns about lack of originality or incremental contribution.

This category includes:
  - Similar ideas already exist in previous work
  - The novelty claims are unclear or overstated
  - The approach appears to be a straightforward extension of existing methods

Example points:
  - "Could the authors clarify what is novel compared to [prior work]?"
  - "How does the proposed method differ from [prior work] in terms of key contributions?"

[Soundness]
Definition: Questions or concerns about errors, flaws, or lack of rigor in the METHOD or theory.

This category includes:
  - Proof gaps or incorrect derivations
  - Assumptions are too strong or unrealistic
  - Claims are unjustified or logically inconsistent
  - Theoretical results are questionable
  - The claimed necessity/superiority of a component is not justified by analysis or evidence

Example points:
  - "What justifies Assumption X, and when might it fail?"
  - "Could the authors clarify why the claim C follows from the analysis?"
  - "What is the justification for stating that component X is indispensable?"

[Experiments]
Definition: Requests/concerns about the EXPERIMENTAL SETUP AND RESULTS, not the method itself.

This category includes:
  - Missing baselines, unfair comparisons, limited datasets
  - Missing ablations or evaluations
  - Missing qualitative results/visualizations
  - Reporting requests: runtime numbers, FLOPs, memory, wall-clock, throughput, scaling plots
  - Evaluation concerns: insufficient metrics, questionable metrics, missing analysis across metrics

Example points:
  - "Could the authors compare against [baseline]?"
  - "Could the authors include an ablation study on component X?"
  - "Could the authors report runtime and memory usage?"
  - "Could the authors provide details of the dataset generation or evaluation setup?"

NOT this category (common mistakes):
  - "The method is not effective" -> Soundness
  - "The method is too slow for practical use" -> Significance

[Clarity]
Definition: Requests/concerns about the paper's presentation and readability (how clearly the paper communicates), not missing experiments or reproducibility details.

This category includes:
  - Ambiguous or confusing statements and wording
  - Unclear definitions or notation usage
  - Unclear figures/tables/legends due to presentation
  - Poor organization that makes the paper hard to follow

Example points:
  - "Could the authors clarify what qualifies the proposed model as a foundation model?"
  - "In Fig. X, what do the colors/rows represent?"
  - "Could the authors clarify the meaning of the statement \"...\"?"

NOT this category:
  - Missing baselines/ablations/results/metrics/runtime numbers -> Experiments
  - Missing hyperparameters/training setup/implementation details/code -> Reproducibility
  - Requests to add more metrics/experiments -> Experiments

[Significance]
Definition: Questions or concerns about limited impact, importance, or practical value of the work.

This category includes:
  - Limited real-world applicability or practical benefit
  - The scope is too narrow
  - Scalability or deployment usefulness is questionable

Example points:
  - "Can the proposed approach be deployed as-is, and what are its practical limitations?"
  - "How might this method transfer to real-world settings or other tasks?"

[Reproducibility]
Definition: Requests/concerns about missing code, insufficient details, or reproducibility issues.

This category includes:
  - Code is not provided
  - Hyperparameters or hyperparameter selection criteria are missing
  - Implementation details are missing
  - Training setup is not specified
  - Dataset splits, preprocessing, or evaluation protocol details are missing

Example points:
  - "Will the authors release the code and configurations?"
  - "How were the hyperparameters chosen?"
  - "What is the training setup (epochs, batch size, optimizer, learning rate schedule)?"
  - "How are the dataset split and evaluation protocol defined?"

[Related Work]
Definition: Requests/concerns about missing citations or incomplete coverage of prior work.

This category includes:
  - Important references are missing
  - Related work is not adequately discussed
  - The paper is not well-positioned in literature

Example points:
  - "Could the authors cite and discuss [important work]?"
  - "How does this work relate to [prior work], and what are the key differences?"

[Other]
Definition: Any other issues not covered above.

Example points:
  - "Are there any criteria?"

## EXAMPLE

Input Text:
"1. Could the authors provide runtime and memory usage?\n2. What is the difference between the proposed method and Prior Work 2023?\n3. It is unclear how hyperparameters were chosen."

Output JSON:
{
  "points": [
    {
      "key_point": "Could the authors provide runtime and memory usage?",
      "category": "Experiments"
    },
    {
      "key_point": "What is the difference between the proposed method and Prior Work 2023?",
      "category": "Related Work"
    },
    {
      "key_point": "How were the hyperparameters chosen?",
      "category": "Reproducibility"
    }
  ]
}

## OUTPUT FORMAT (MUST FOLLOW)
Return ONLY one JSON object with root key "points".
Each point has "key_point" (string) and "category" (string).
The output must be valid JSON with double quotes.
Do NOT add any explanations or markdown code blocks.
If no points found, return {"points": []}
"""
\end{lstlisting}
\end{tcolorbox}

\subsection{Point Matching Prompts}
\label{app:matching_prompts}

To determine whether an AI-generated point matches a human point, we use a semantic matching function implemented via LLM-as-a-Judge. The matching prompt evaluates semantic similarity and argument alignment.

\begin{tcolorbox}[
    colback=cyan!10!white,
    colframe=cyan!70!blue,
    title=\textbf{Point Matching Prompts},
    fonttitle=\bfseries,
    breakable,
    boxrule=1pt,
    arc=3pt,
    left=8pt,
    right=8pt,
    top=8pt,
    bottom=8pt
]
\begin{lstlisting}[
    basicstyle=\small\ttfamily,
    breaklines=true,
    breakatwhitespace=true,
    breakindent=0pt,
    columns=fullflexible,
    keepspaces=true,
    showstringspaces=false,
    showtabs=false,
    tabsize=2,
    aboveskip=0pt,
    belowskip=0pt
]
"""The following are two review points from paper peer reviews.
Determine whether they describe the same underlying point or claim.

Point A: {point_a}

Point B: {point_b}

Rules:
1. Focus on the core meaning and specific claim, ignore wording differences.
2. Answer "yes" only if they describe the same point; loosely related or overlapping topics are "no".

Respond in JSON only: {{"match":"yes"}} or {{"match":"no"}}."""
\end{lstlisting}
\end{tcolorbox}

\subsection{Question Evaluation Prompts}
\label{app:question_eval_prompts}

For evaluating the \textit{question} field, we employ prompts to assess both confidence and constructiveness of each question point.

\begin{tcolorbox}[
    colback=cyan!10!white,
    colframe=cyan!70!blue,
    title=\textbf{Question Type Prompts},
    fonttitle=\bfseries,
    breakable,
    boxrule=1pt,
    arc=3pt,
    left=8pt,
    right=8pt,
    top=8pt,
    bottom=8pt
]
\begin{lstlisting}[
    basicstyle=\small\ttfamily,
    breaklines=true,
    breakatwhitespace=true,
    breakindent=0pt,
    columns=fullflexible,
    keepspaces=true,
    showstringspaces=false,
    showtabs=false,
    tabsize=2,
    aboveskip=0pt,
    belowskip=0pt
]
"""
## Role
You are an expert scientific manuscript editor. Your task is to determine the underlying intent of a peer review question.

## Classification Logic
Evaluate the question based on the following behavioral intents:

- **explain**: The question asks the author to further explain, clarify, or provide evidence for **some content**.
  *Examples (including but not limited to)*: Asking "why", "what", or "how"; requesting clarification of a definition; explaining the motivation behind a choice; or detailing an existing mathematical derivation or experimental setup.

- **supplement**: The question requests **supporting materials**, additional experiments, or substantial modifications.
  *Examples (including but not limited to)*: Requesting additional baseline comparisons; performing new ablation studies; providing extra code/data; or asking for changes to the model architecture or methodology.

- **other**: The question does not fit into the above categories.
  *Examples (including but not limited to)*: Purely formatting issues; grammatical corrections; citation suggestions; or general comments without specific technical requests (e.g., broad praise or vague criticism).

## Constraints
- **Return JSON only.** No preamble, no postscript, no explanation.
- Identify the most dominant intent. 

Question: {question}

## Output Format
{{"type": "explain|supplement|other"}}"""
\end{lstlisting}
\end{tcolorbox}

\begin{tcolorbox}[
    colback=cyan!10!white,
    colframe=cyan!70!blue,
    title=\textbf{Explain Eval Prompts},
    fonttitle=\bfseries,
    breakable,
    boxrule=1pt,
    arc=3pt,
    left=8pt,
    right=8pt,
    top=8pt,
    bottom=8pt
]
\begin{lstlisting}[
    basicstyle=\small\ttfamily,
    breaklines=true,
    breakatwhitespace=true,
    breakindent=0pt,
    columns=fullflexible,
    keepspaces=true,
    showstringspaces=false,
    showtabs=false,
    tabsize=2,
    aboveskip=0pt,
    belowskip=0pt
]
"""
# Task Background
We are checking if the reviewer's question is grounded in the paper. Our objective is to conduct a grounding audit to verify whether the specific explanation, reason, or technical detail requested by the reviewer has already appeared or been addressed in this paper chunk. We want to determine if the reviewer is asking for information that the authors have already provided, thereby assessing the quality and factual basis of the review.

# Role
You are an expert scientific manuscript editor. Your task is to decide whether the provided paper chunk explicitly answers the question or provides the justification sought by the reviewer.

# Evaluation Rules
- **Assign "yes" ONLY if**:
  1. The chunk explicitly provides the reasoning, definition, or precise technical detail the question asks for.
  2. The chunk provides the logical motivation or evidence that satisfies the "why" or "how" of the inquiry.
  3. The explanation is present through semantic sufficiency; if the core meaning is addressed but phrased differently, assign "yes".

- **Assign "no" if**:
  1. The chunk only mentions the concept or topic without providing the required explanation or depth.
  2. The chunk introduces the subject but does not resolve the specific technical doubt raised by the reviewer.
  3. The information is too vague or requires excessive inference from the reader to count as a direct answer.

# Input Data
- **Question**: {question}
- **Paper Chunk**: {chunk}

# Constraints
- Return JSON only. No preamble or postscript.
- Be highly critical. Do not give a "yes" for keyword matches alone; the explanatory intent must be satisfied.

# Output Format
{{"has_info": "yes|no"}}
"""
\end{lstlisting}
\end{tcolorbox}

\begin{tcolorbox}[
    colback=cyan!10!white,
    colframe=cyan!70!blue,
    title=\textbf{Supplement Eval Prompts},
    fonttitle=\bfseries,
    breakable,
    boxrule=1pt,
    arc=3pt,
    left=8pt,
    right=8pt,
    top=8pt,
    bottom=8pt
]
\begin{lstlisting}[
    basicstyle=\small\ttfamily,
    breaklines=true,
    breakatwhitespace=true,
    breakindent=0pt,
    columns=fullflexible,
    keepspaces=true,
    showstringspaces=false,
    showtabs=false,
    tabsize=2,
    aboveskip=0pt,
    belowskip=0pt
]
"""
# Task Background
We are evaluating the constructiveness of a reviewer's suggestion. We need to check if the requested new work (e.g., experiments, data, or changes) has already appeared in this paper chunk. If it has, the reviewer's suggestion is considered redundant.

# Role
You are an expert scientific manuscript editor. Decide whether the requested new work is already present in this chunk.

# Evaluation Rules
- Assign "yes" ONLY if the requested supplement (e.g., additional experiment, ablation, baseline, or modification) is ALREADY featured in this chunk.
- Assign "no" if the work is truly missing or only briefly hinted at without substantive results.

# Input Data
- **Question**: {question}
- **Paper Chunk**: {chunk}

# Constraints
- Return JSON only. No preamble or postscript.
- Use high rigor. Only assign "yes" if the evidence of the completed work is clear and direct.

# Output Format
{{"has_info": "yes|no"}}
"""
\end{lstlisting}
\end{tcolorbox}

\subsection{Prompt Variations and Experiments}
\label{app:prompt_variations}

We conducted extensive prompt debugging experiments to understand the impact of prompt design on model outputs. Our experiments revealed that:

\begin{itemize}
    \item Rating outputs are highly susceptible to prompt influence
    \item Neutral prompts tend to produce overly lenient rating distributions.Under neutral prompt conditions, predicted ratings for both Qwen and Llama series are predominantly clustered within the (8, 10) range, significantly undermining inference effectiveness.
    \item A bit strict prompts help maintain more balanced and critical evaluation patterns
    \item Prompt engineering can significantly affect MAE without necessarily improving review quality
\end{itemize}

\section{LLM-Generated Review Examples}
\label{app:llm_review_examples}

This section provides a representative example of reviews generated by LLMs in our evaluation framework. The example illustrate the quality, style, and characteristics of agent-generated reviews.

\begin{tcolorbox}[
    colback=pink!20!white,
    colframe=pink!60!red,
    title=\textbf{Agent-generated Review},
    fonttitle=\bfseries,
    breakable,
    boxrule=1pt,
    arc=3pt,
    left=8pt,
    right=8pt,
    top=8pt,
    bottom=8pt
]
\begin{lstlisting}[
    basicstyle=\small\ttfamily,
    breaklines=true,
    breakatwhitespace=true,
    breakindent=0pt,
    columns=fullflexible,
    keepspaces=true,
    showstringspaces=false,
    showtabs=false,
    tabsize=2,
    aboveskip=0pt,
    belowskip=0pt
]
{
  "summary": "This paper introduces Mask-based Latent Reconstruction (MLR), a self-supervised method for improving state representation learning in reinforcement learning (RL) by reconstructing masked pixels in the latent space. The method integrates with RL training as an auxiliary objective, enabling better coordination between representation learning and policy learning. The authors conduct extensive experiments on continuous and discrete control benchmarks, showing that MLR improves sample efficiency and outperforms existing methods. The paper also explores different masking strategies and reconstruction targets, and provides ablation studies to validate the effectiveness of their approach.",
  "strengths": "The paper presents a novel approach to state representation learning in RL by leveraging mask-based reconstruction in the latent space, which is a creative adaptation of techniques from NLP and CV. The method is well-motivated, with a clear connection to prior work on self-supervised learning. The experiments are comprehensive, covering multiple benchmarks and ablation studies, which support the claims of improved sample efficiency. The paper also provides a detailed analysis of different design choices, such as masking strategies and reconstruction targets, which adds depth to the work. The integration of MLR as an auxiliary objective during RL training is a significant contribution, as it avoids the need for additional rollouts and aligns with the goal of sample efficiency.",
  "weaknesses": "The paper lacks a thorough comparison with other self-supervised methods in RL, such as contrastive learning or other auxiliary tasks like reward prediction. The experimental results, while promising, are presented with limited statistical analysis, particularly in the context of high-variance benchmarks like Atari. The paper also does not clearly explain how the proposed method compares to the state-of-the-art in terms of computational efficiency or training time. Additionally, the theoretical justification for why latent space reconstruction is more effective than pixel-level reconstruction is not sufficiently developed. The paper could benefit from a more rigorous analysis of the trade-offs between different masking strategies and their impact on policy learning.",
  "questions": "1. How does MLR compare to other self-supervised methods in RL, such as contrastive learning or reward prediction? 2. What is the exact impact of the choice of mask ratio on the performance of MLR, and are there any optimal values beyond 50% that could be explored? 3. The paper mentions that the latent space reconstruction is more effective than pixel-level reconstruction, but what is the theoretical basis for this claim? 4. How does the computational cost of MLR compare to existing methods in terms of training time and resource usage? 5. Are there any limitations or scenarios where MLR might not be effective, and how does the paper address these?",
  "soundness": 3,
  "presentation": 3,
  "contribution": 3,
  "rating": 5,
  "confidence": 4,
  "paper_id": "-zlJOVc580",
  "model_name": "Qwen3-8B"
}
\end{lstlisting}
\end{tcolorbox}

\section{Point Extraction Categories}
\label{app:point_categories}

We classify extracted argument points into 8 distinct categories. Below are detailed definitions and examples for each category:

\begin{enumerate}
    \item \textbf{Novelty:} Focuses on creativity and originality of the research contribution. This category captures assessments of whether the work introduces new ideas, methods, or perspectives that advance the field.
    
    \item \textbf{Soundness:} Evaluates the correctness of methodology and theoretical proofs. This includes assessments of mathematical rigor, logical consistency, and methodological validity. Note that "method effectiveness" belongs to Soundness, while "good experimental results" belongs to Experiments.
    
    \item \textbf{Experiments:} Covers experimental design and result data. This category includes evaluations of experimental setup, data quality, statistical analysis, and result interpretation. Distinguishing between main experiments and ablation experiments can be challenging without full paper context.
    
    \item \textbf{Clarity:} Evaluates writing quality and figure presentation. This includes assessments of paper organization, writing clarity, figure quality, and overall presentation effectiveness.
    
    \item \textbf{Significance:} Assesses practical value and impact of the research. This category captures evaluations of the work's importance, potential applications, and contribution to the field.
    
    \item \textbf{Reproducibility:} Focuses on the completeness of code and parameters. This includes assessments of whether sufficient information is provided for reproducing the results, code availability, and parameter documentation.
    
    \item \textbf{Related Work:} Evaluates the sufficiency of literature citations. This category includes assessments of whether relevant prior work is properly cited and discussed, and whether the work is properly positioned within the existing literature.
    
    \item \textbf{Other:} Includes additional considerations such as ethics, societal impact, and other factors that do not fit into the above categories.
\end{enumerate}

\section{Dataset Construction and Refinement}
\label{sec:dataset}

To support the proposed evaluation framework and facilitate robust instruction tuning, we construct a large-scale, high-quality dataset of scientific peer reviews. We collect data from OpenReview, covering NeurIPS (2022--2025) and ICLR (2024--2026), totaling 46,199 papers and their corresponding review data. The raw data presents significant heterogeneity in scoring scales, text structures, and quality. To construct a high-quality test set, we apply a multi-stage pipeline to standardize, filter, clean, and enrich the raw datasets.

\subsection{Data Standardization}
\textbf{Unified Rating Schema.}
Different conferences and years employ varying scoring scales. For example, NeurIPS 2022--2024 all use a 10-point scale, while NeurIPS 2025 uses a 6-point scale. ICLR 2026 also differs from ICLR 2024 and 2025 in scoring scales.

To address this issue and achieve cross-conference and cross-year rating consistency, we map all ratings to the ICLR standard scale $\mathcal{S} = \{1, 3, 5, 6, 8, 10\}$ according to the reviewing standards of each conference. This mapping is performed according to the specific reviewing guidelines of each conference year to preserve the semantic meaning of "Accept", "Weak Accept", etc.

\textbf{Structural Alignment.}
While ICLR conference data explicitly separates \textit{Strengths} and \textit{Weaknesses}, NeurIPS conference in 2022 and 2025 combines them into a single field \textit{strength\_and\_weakness}.

To address this issue and align the data structure, we employ a LLM (Qwen3-235B) to semantically parse the \textit{strength\_and\_weakness} field in these data and decouple them into independent fields \textit{strengths} and \textit{weaknesses} to ensure structural consistency.

\textbf{Metadata Cleaning.}
The \textit{Decision} fields across different conferences and years exhibit inconsistent specifications. For example, capitalization is inconsistent, and some data contain messy HTML tags or special characters.

To address this issue and unify data specifications, we normalize the \textit{Decision} fields using regular expressions to clean the data and remove garbage data.

Notably, NeurIPS conference data exhibits a survival bias in the \textit{Decision} field, as rejected papers are typically not public (approximately 95\% acceptance), and we acknowledge this distribution shift.

\subsection{Quality Filtering}
The raw datasets obtained from OpenReview contain review data with significant quality variations. To address this issue, we conduct detailed statistical analysis on the review data. Based on the statistical information and data quality requirements, we design and apply strict filtering criteria to clean disordered data, filter high-quality samples to construct the dataset to better conduct subsequent experimental tests.

\begin{itemize}
    \item \textbf{Expertise Filter:} To select high-quality review data, we focus on the \textit{Confidence} field, which reflects the reviewer's confidence level in the review content, and reviews with high confidence are typically of higher quality. Therefore, we retain reviews where the reviewer's self-assessed confidence is high, with a confidence threshold of: $\text{Confidence} \in \{4, 5\}$.
    \item \textbf{Review Count Constraint:} Statistical analysis reveals that after removing low-confidence reviews, some papers have insufficient review counts, lacking validity in review data. To address this, we filter out papers that do not maintain a reasonable number of valid reviews, with a review count threshold of: $N \in \{3, 4, 5\}$.
    \item \textbf{Consensus Filtering:} For papers with significant controversy, their rating fields often have large variance, which introduces excessive noise during training. To avoid training on ambiguous signals, we aim to remove papers with high controversy. By calculating the variance of ratings for each paper and analyzing the statistical data curve, we find that a variance of 1.5 is the inflection point that balances data volume and label consistency. We design a rating threshold of: $\sigma^2 \le 1.5$.
\end{itemize}
After the above data cleaning process, we obtain a refined dataset of over 16,000 papers.
 
\subsection{Granular Annotation: Point Extraction}
A key contribution of our proposed dataset is the fine-grained annotation of review texts, which enables the \textit{Argument Quality} evaluation described in Section~\ref{sec:evaluation}.

\textbf{Atomic Point Extraction.} We use the Qwen3-235B model for inference to decompose the raw text of \textit{Strengths} and \textit{Weaknesses} into atomic points. The extraction process adheres to five rules: \textit{Split Compounds}, \textit{Causal Decomposition}, \textit{Preserve Integrity}, \textit{Coreference Resolution}, \textit{Noise Removal}.

\textbf{Taxonomy and Classification.} We classify the extracted points into 8 categories and add two list fields \textit{strength\_points} and \textit{weakness\_points} to the data for storage. These 8 categories are as follows: \textit{Novelty}, \textit{Soundness}, \textit{Experiments}, \textit{Clarity}, \textit{Significance}, \textit{Reproducibility}, \textit{Related Work}, \textit{Other}.
Notably, for main experiments and ablation experiments, we observe that LLMs struggle to distinguish these reliably without access to the full paper content. Therefore, we refined our prompt to enforce a strict definition of the general \textit{Experiments} category, ensuring high classification accuracy.

\section{Binoculars Algorithm}
\label{app:binoculars}
\textbf{Binoculars Method Principle.} The Binoculars method\citep{hans2024spotting} is based on a dual-model comparison mechanism: the observer model $M_1$ reads the input text and calculates its perplexity. If the text closely conforms to machine statistical patterns (common vocabulary, standard sentence structures), the perplexity is low; if the text contains human-specific jump thinking, rare expressions, or unique styles, the perplexity is typically higher. Meanwhile, the baseline model $M_2$ makes predictions for the same context (simulating machine generation behavior), and then the observer model $M_1$ evaluates the perplexity of these machine-predicted words.

primary rationale for selecting Binoculars as our AI-detection tool is its methodological foundation in perplexity-based calculations. Lower Binoculars scores typically signify that the text is composed of formulaic language, suggesting a lack of genuine semantic understanding or intellectual depth in the generated review. From this perspective, the Binoculars framework serves as an implicit proxy for evaluating the overall quality and substantive depth of the content.

\textbf{Score Calculation.} Let $P_{\text{actual}}$ denote the perplexity of the observer model $M_1$ on the actual text, and $P_{\text{baseline}}$ denote the perplexity of $M_1$ on the machine baseline $M_2$ predictions. The Binoculars Score is calculated as:
\begin{equation}
    \text{Binoculars Score} = \frac{\log(P_{\text{actual}})}{\log(P_{\text{baseline}})}
\end{equation}
From a mathematical definition perspective, the theoretical range of Binoculars Score is $(0, \infty)$, but the vast majority of texts have scores falling within the interval of 0.7 to 1.3. Higher scores indicate closer proximity to human writing, while lower scores indicate closer proximity to AI output.

We use this as a quality control metric to prevent the generation of generic, template-like reviews, ensuring that AI-generated reviews have sufficient linguistic diversity and naturalness.

\section{Rating and MAE Analysis Figures}
\label{app:rating_mae_figures}

This section presents comprehensive visualizations of the relationship between rating scores and MAE across different evaluation metrics and review fields. The figures illustrate how various text similarity metrics (ROUGE-1, ROUGE-2, BLEU-2, BLEU-4, and BERTScore) correlate with rating prediction accuracy.

\begin{figure*}[htbp]
    \centering
    \begin{subfigure}[t]{0.48\textwidth}
        \centering
        \includegraphics[width=\linewidth]{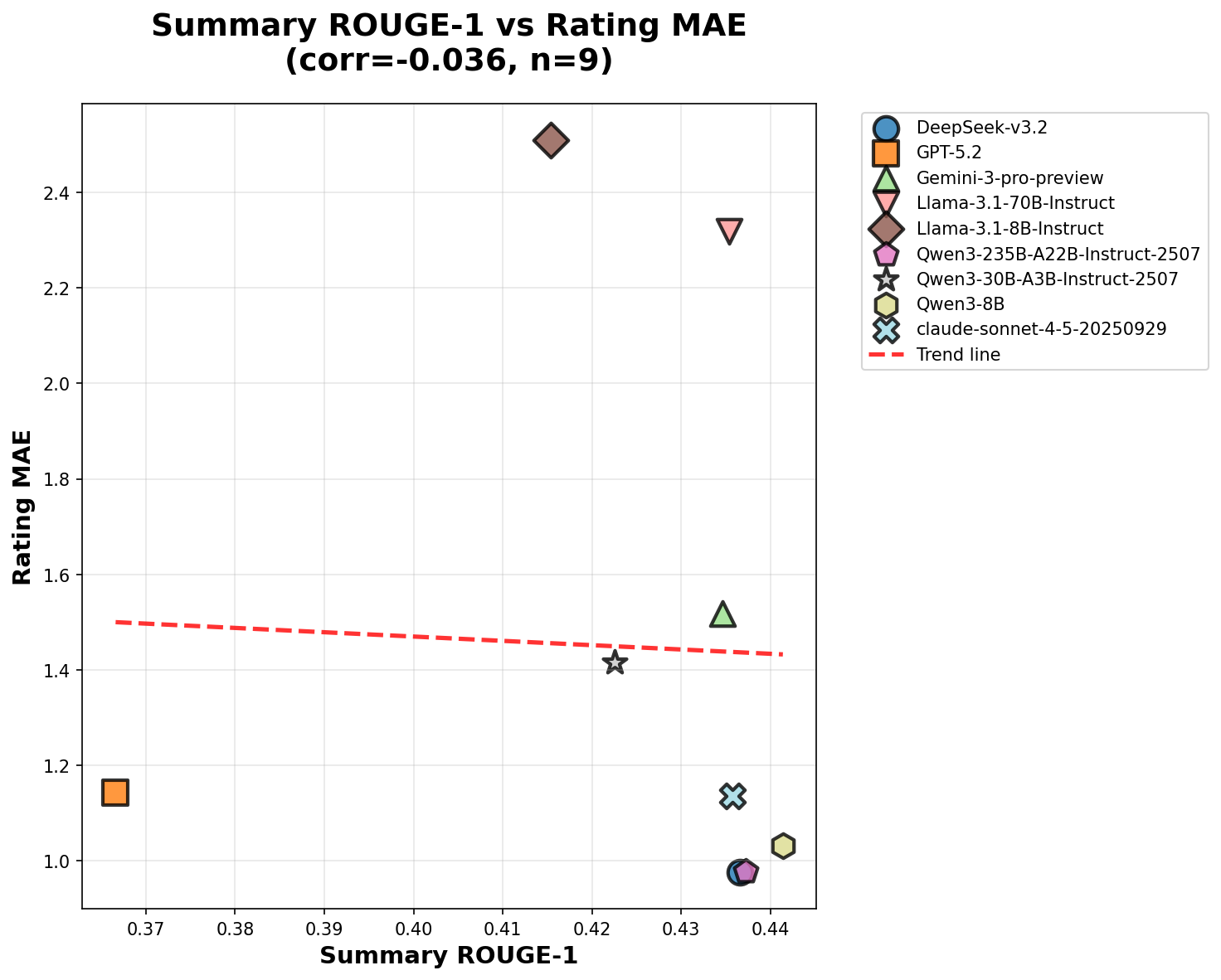}
        \caption{ROUGE-1}
        \label{fig:summary_rouge1}
    \end{subfigure}
    \hfill
    \begin{subfigure}[t]{0.48\textwidth}
        \centering
        \includegraphics[width=\linewidth]{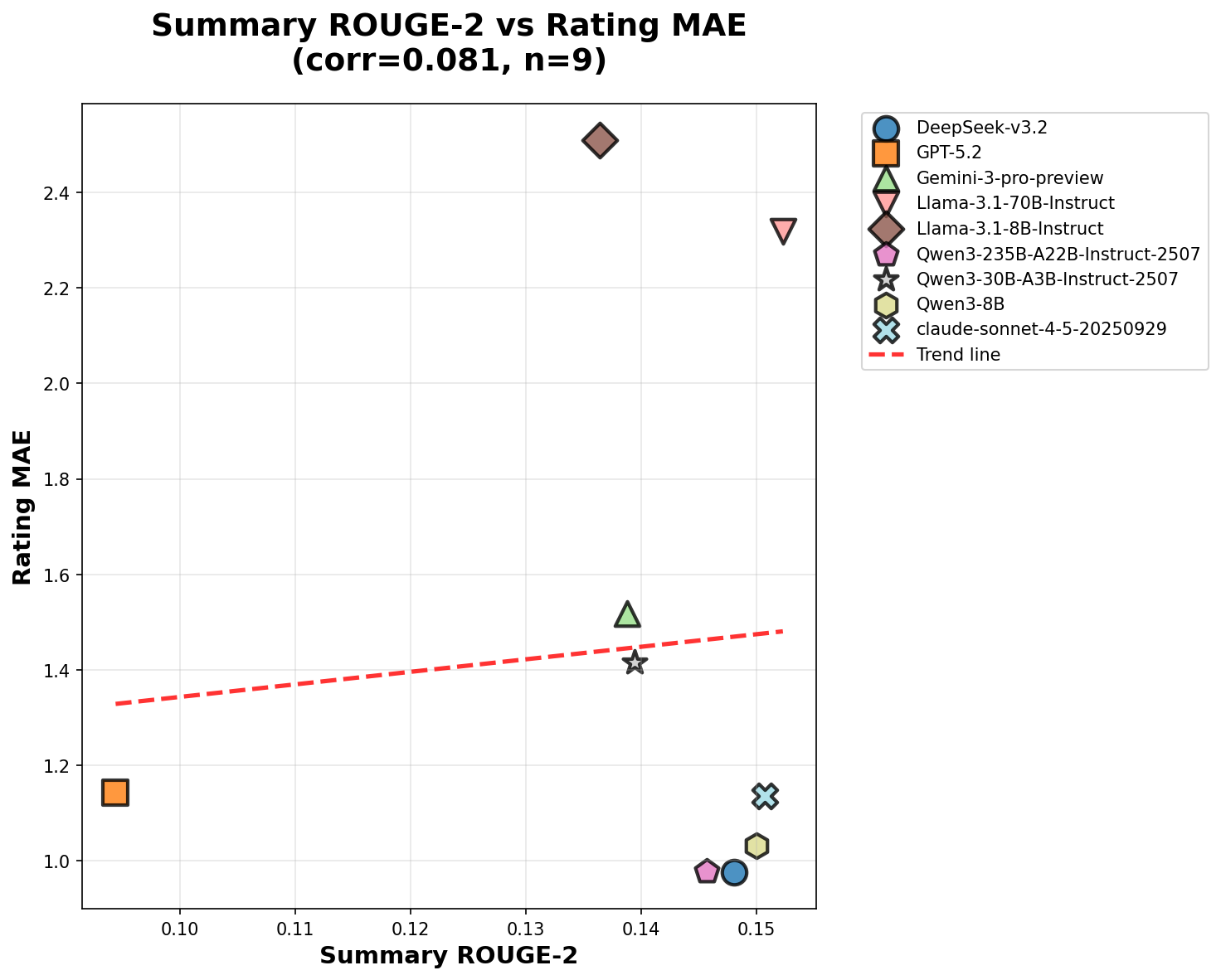}
        \caption{ROUGE-2}
        \label{fig:summary_rouge2}
    \end{subfigure}
    
    \begin{subfigure}[t]{0.48\textwidth}
        \centering
        \includegraphics[width=\linewidth]{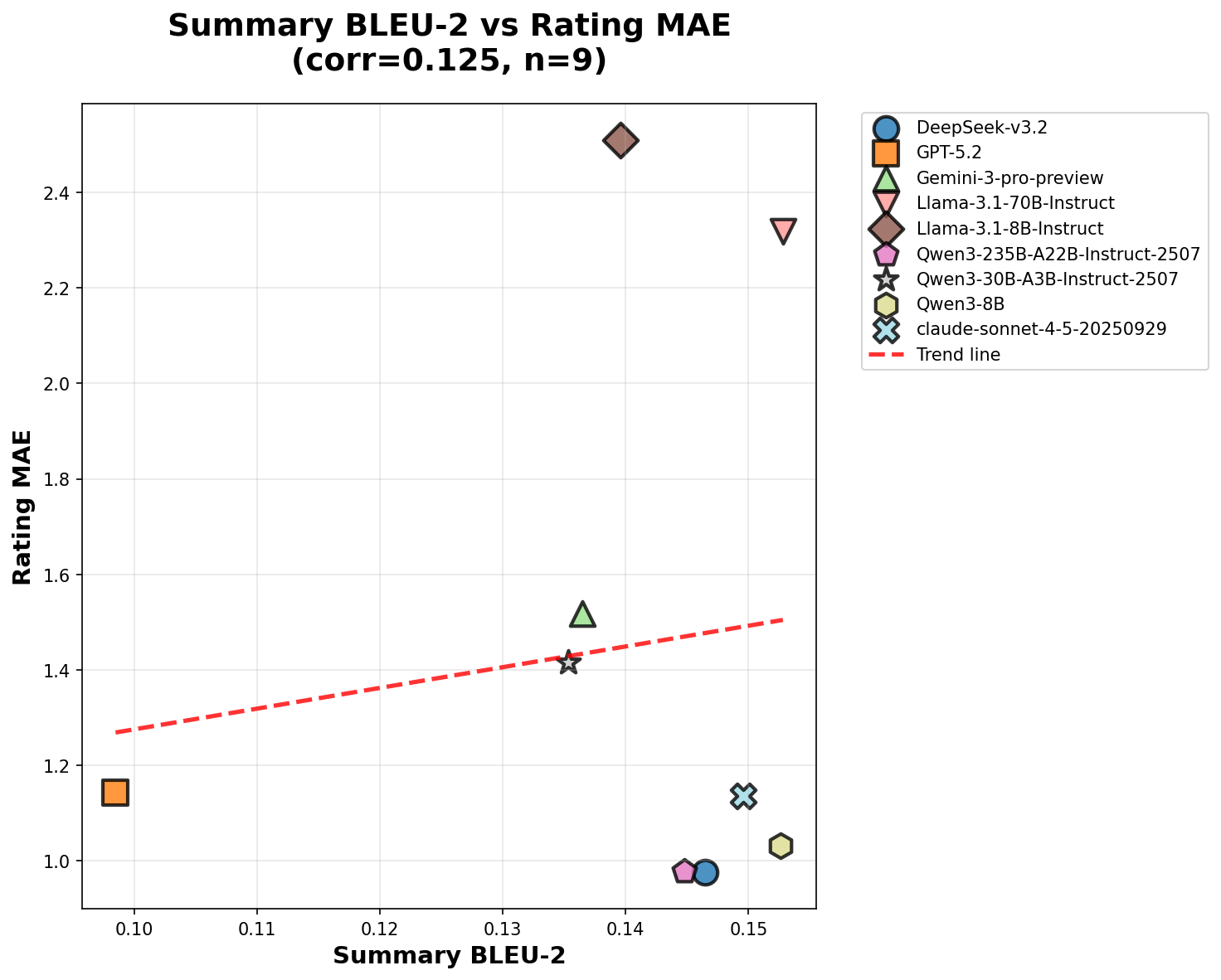}
        \caption{BLEU-2}
        \label{fig:summary_bleu2}
    \end{subfigure}
    \hfill
    \begin{subfigure}[t]{0.48\textwidth}
        \centering
        \includegraphics[width=\linewidth]{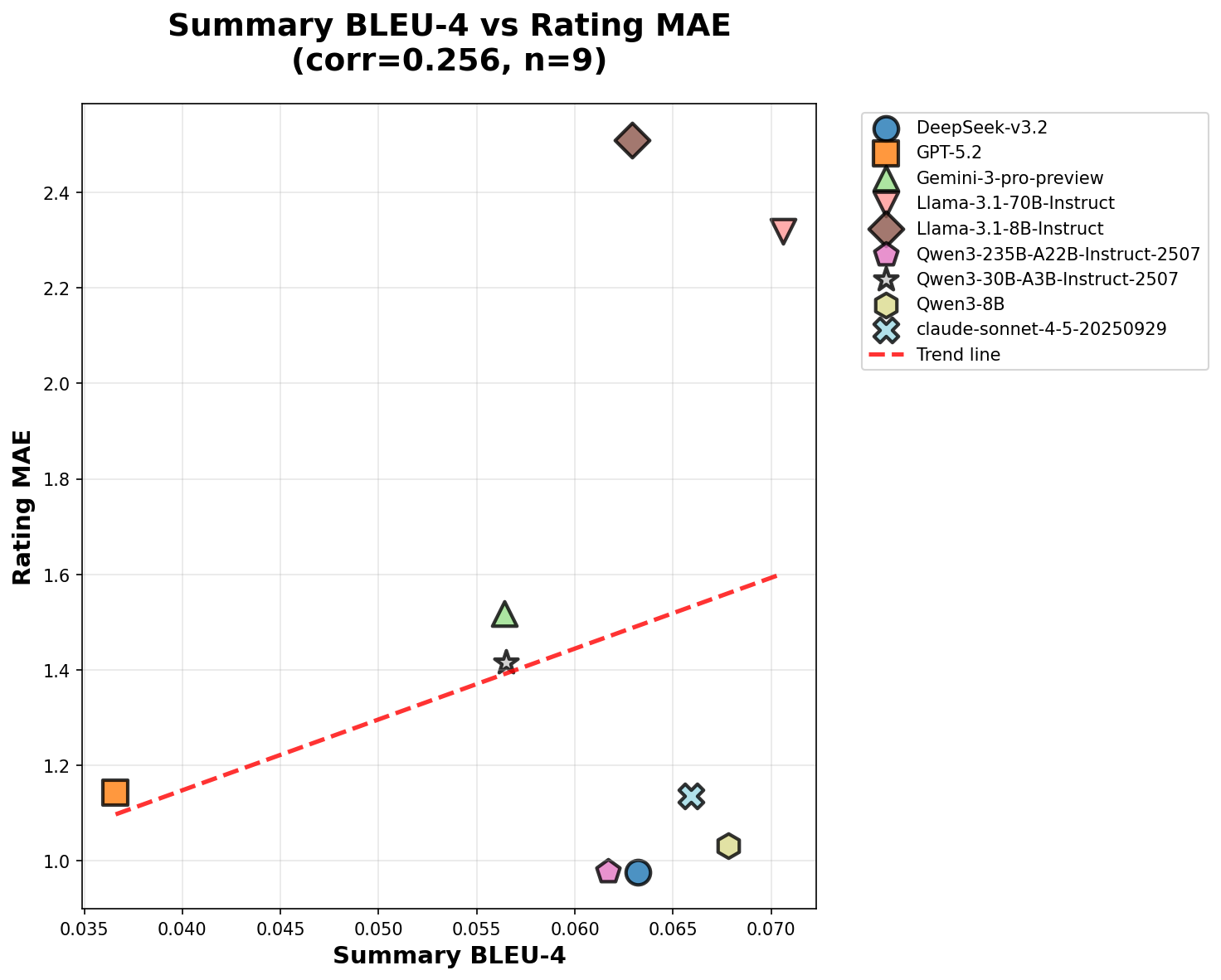}
        \caption{BLEU-4}
        \label{fig:summary_bleu4}
    \end{subfigure}
    
    \begin{subfigure}[t]{0.48\textwidth}
        \centering
        \includegraphics[width=\linewidth]{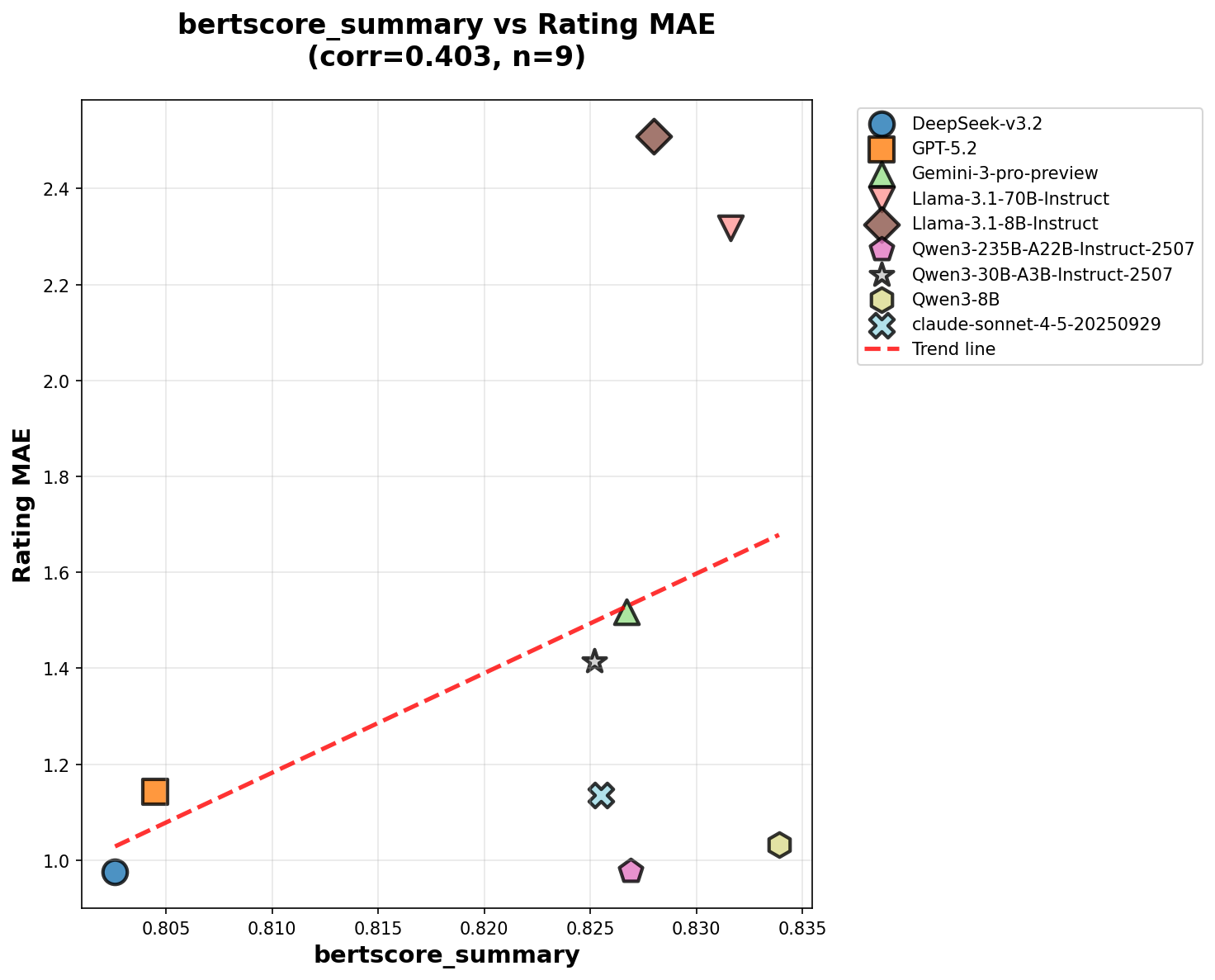}
        \caption{BERTScore}
        \label{fig:bertscore_summary}
    \end{subfigure}
    
    \caption{Rating vs. MAE analysis for the Summary field across different evaluation metrics.}
    \label{fig:summary_rating_mae}
\end{figure*}

\begin{figure*}[htbp]
    \centering
    \begin{subfigure}[t]{0.48\textwidth}
        \centering
        \includegraphics[width=\linewidth]{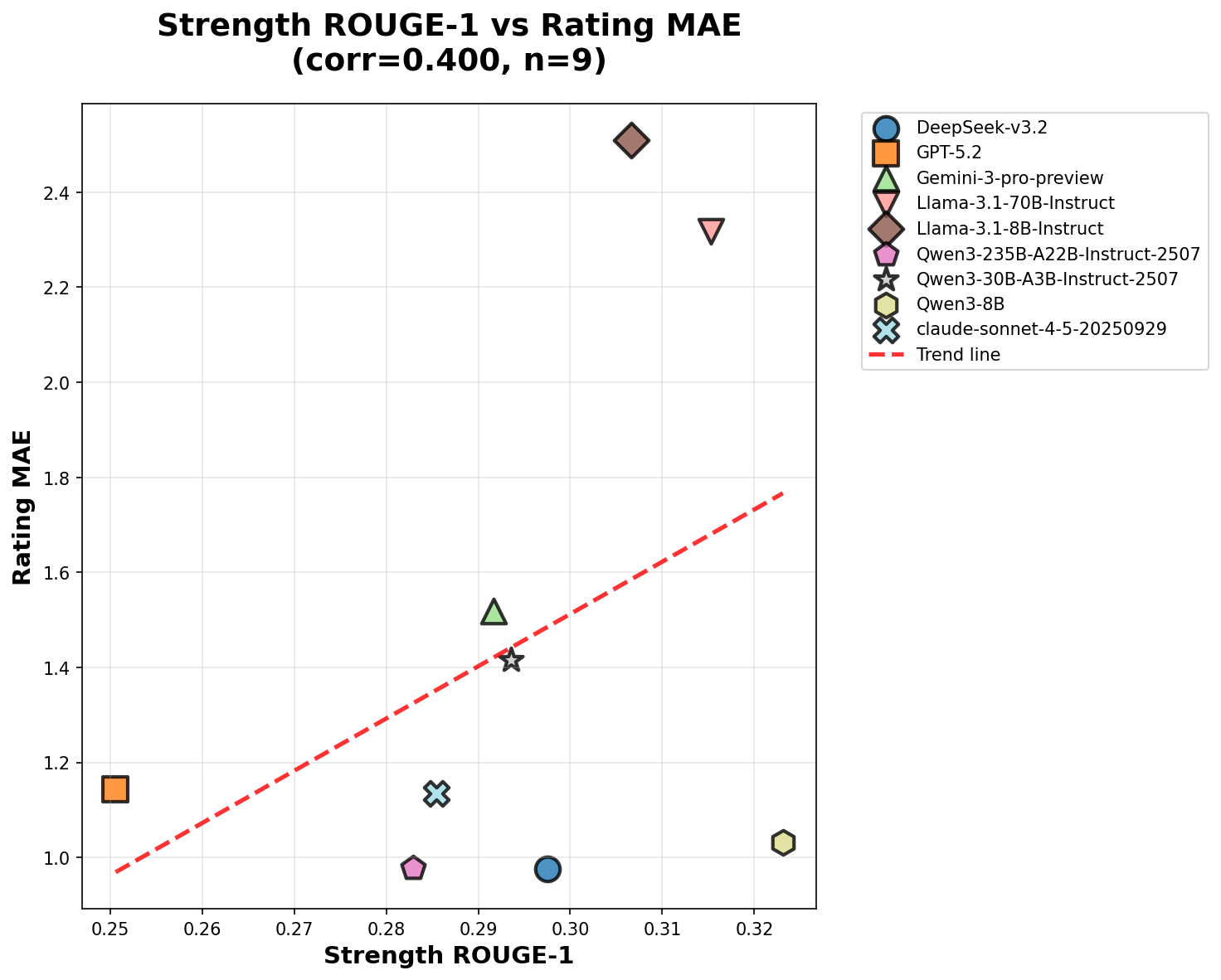}
        \caption{ROUGE-1}
        \label{fig:strength_rouge1}
    \end{subfigure}
    \hfill
    \begin{subfigure}[t]{0.48\textwidth}
        \centering
        \includegraphics[width=\linewidth]{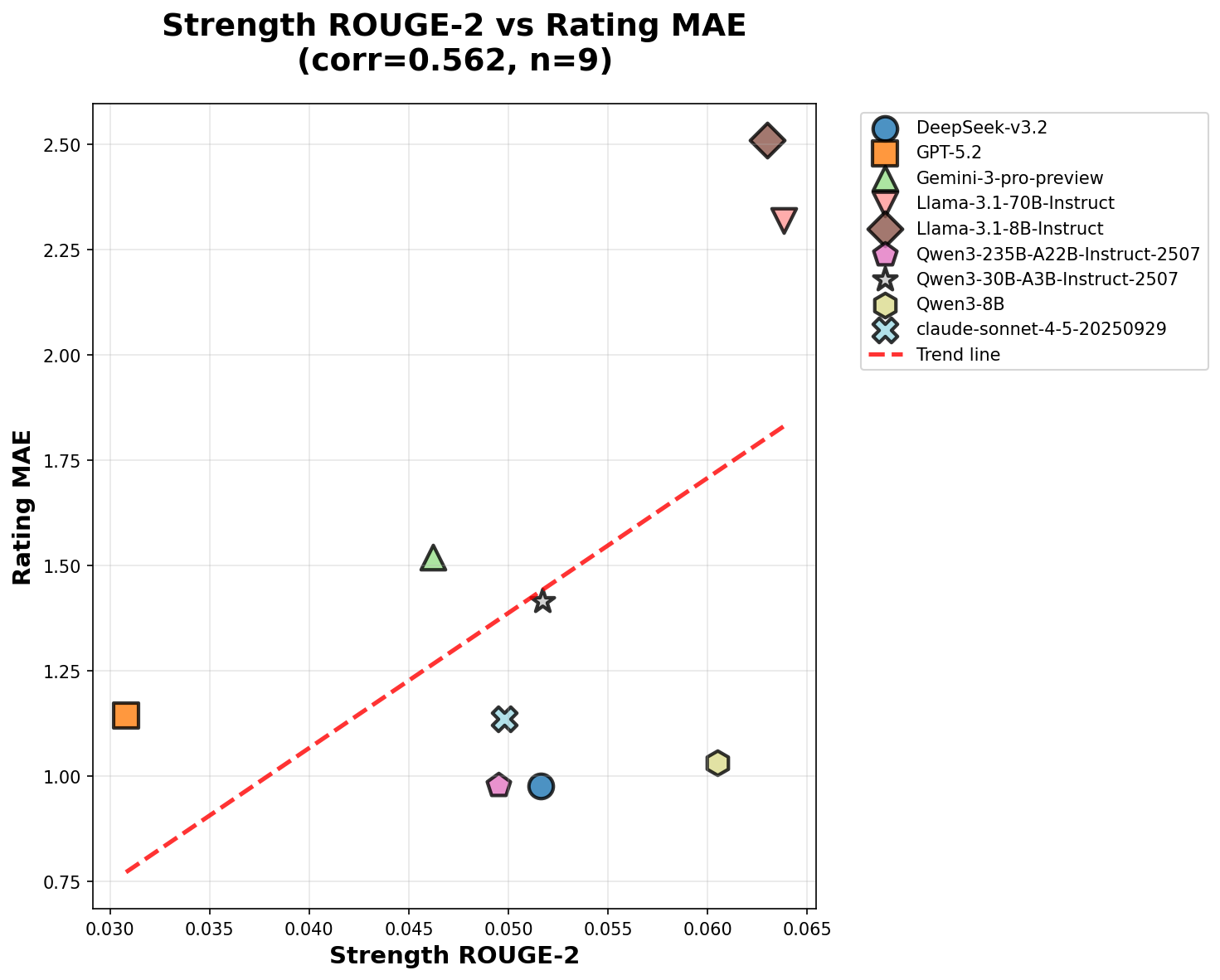}
        \caption{ROUGE-2}
        \label{fig:strength_rouge2}
    \end{subfigure}
    
    \begin{subfigure}[t]{0.48\textwidth}
        \centering
        \includegraphics[width=\linewidth]{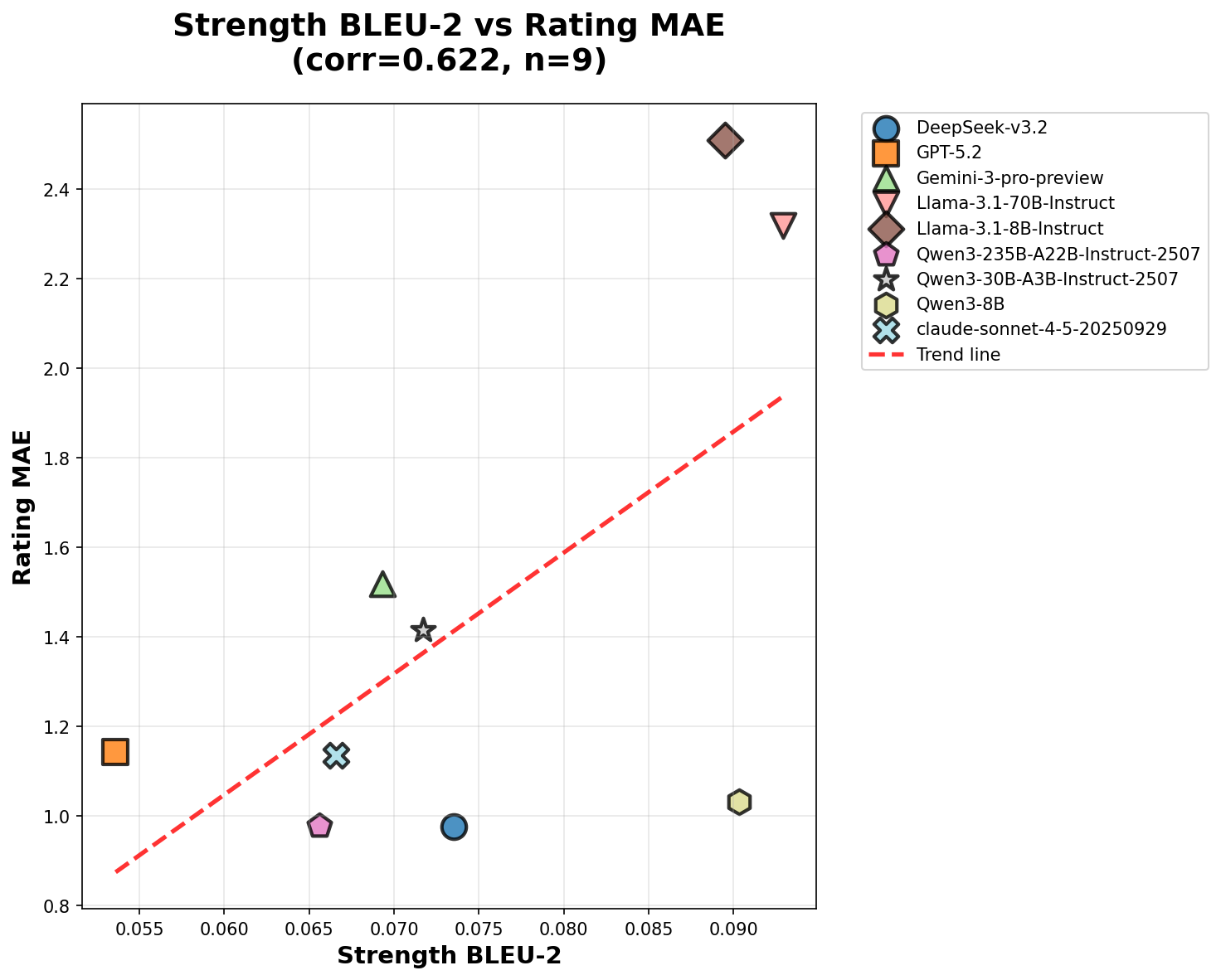}
        \caption{BLEU-2}
        \label{fig:strength_bleu2}
    \end{subfigure}
    \hfill
    \begin{subfigure}[t]{0.48\textwidth}
        \centering
        \includegraphics[width=\linewidth]{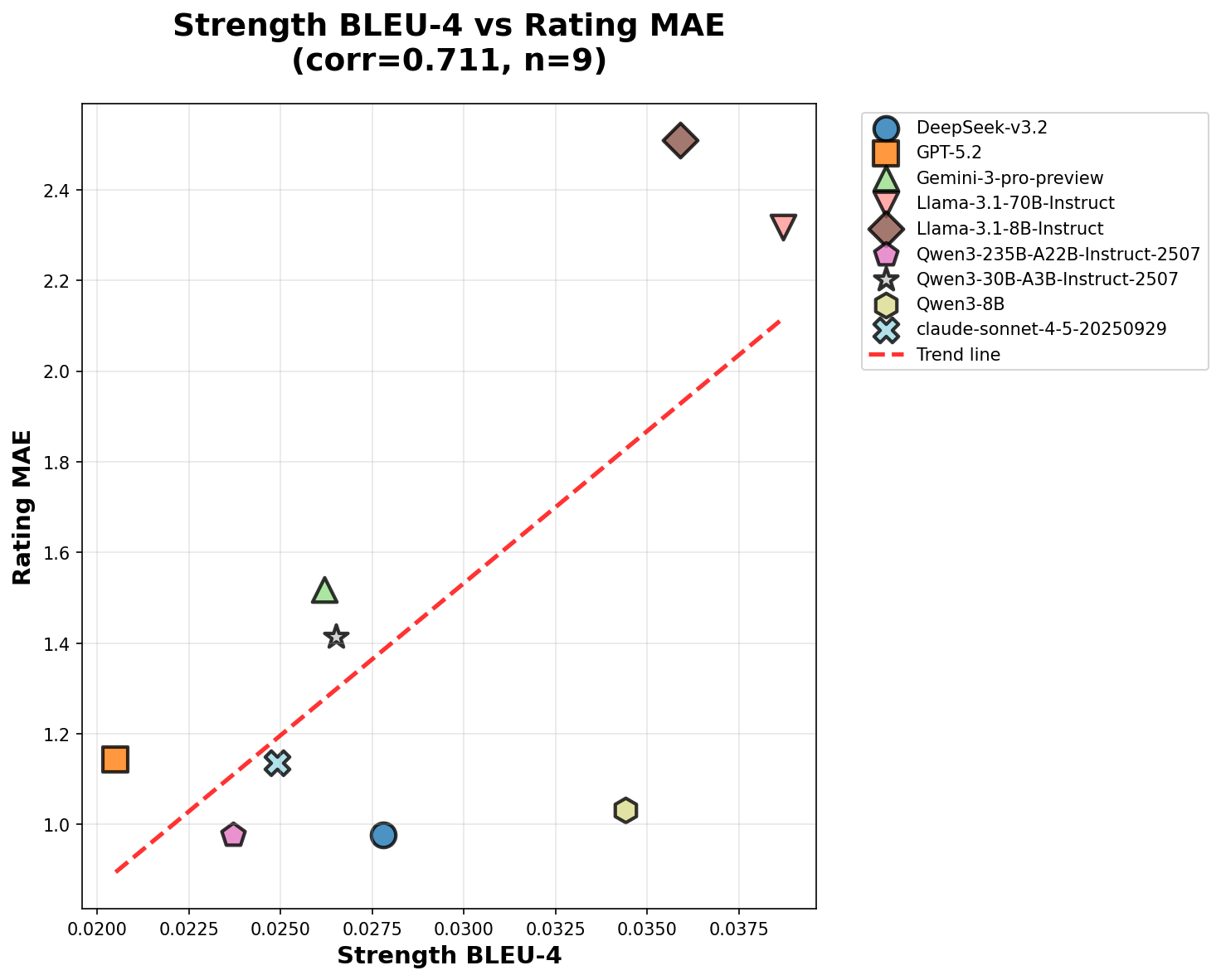}
        \caption{BLEU-4}
        \label{fig:strength_bleu4}
    \end{subfigure}
    
    \begin{subfigure}[t]{0.48\textwidth}
        \centering
        \includegraphics[width=\linewidth]{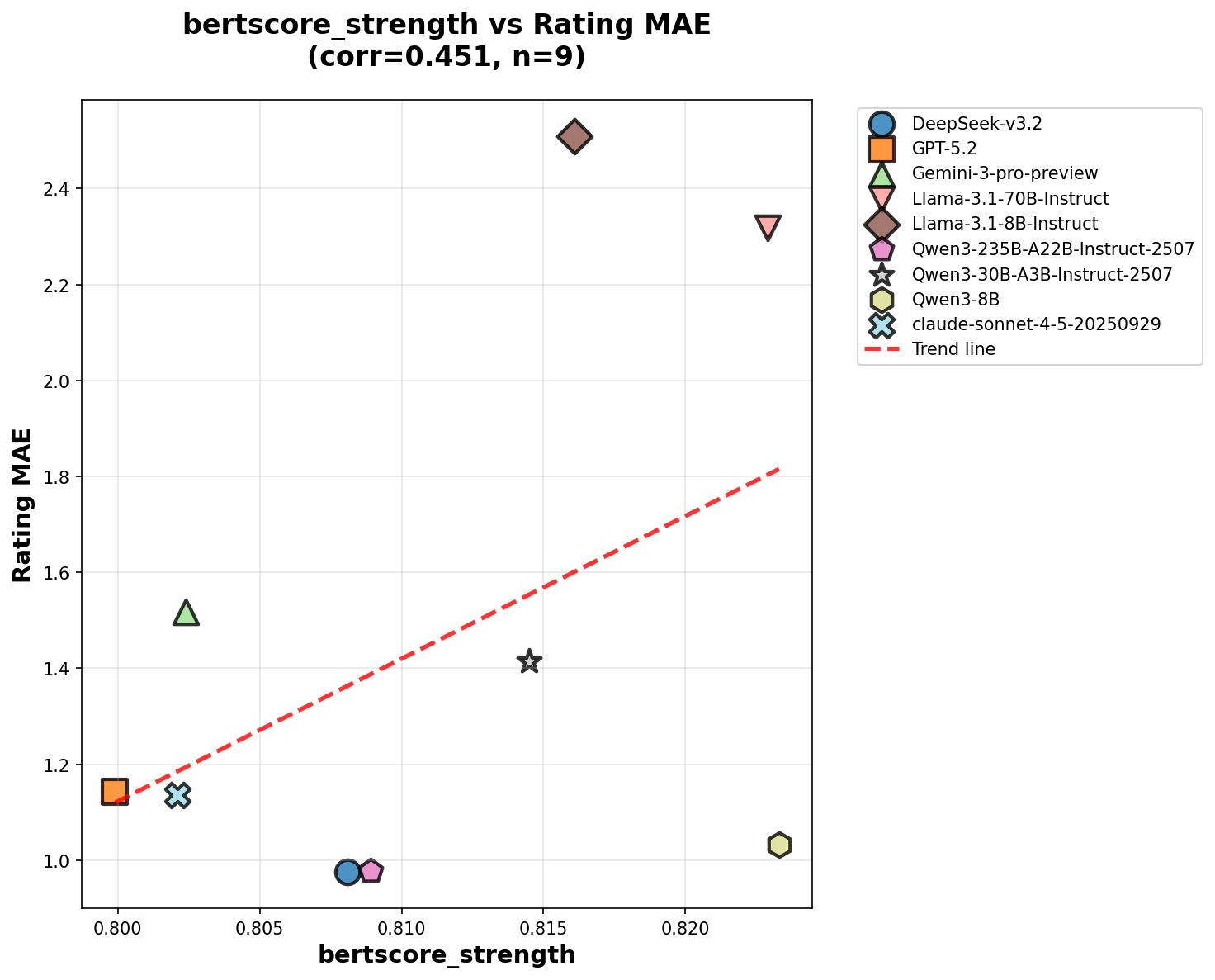}
        \caption{BERTScore}
        \label{fig:bertscore_strength}
    \end{subfigure}
    
    \caption{Rating vs. MAE analysis for the Strength field across different evaluation metrics.}
    \label{fig:strength_rating_mae}
\end{figure*}

\begin{figure*}[htbp]
    \centering
    \begin{subfigure}[t]{0.48\textwidth}
        \centering
        \includegraphics[width=\linewidth]{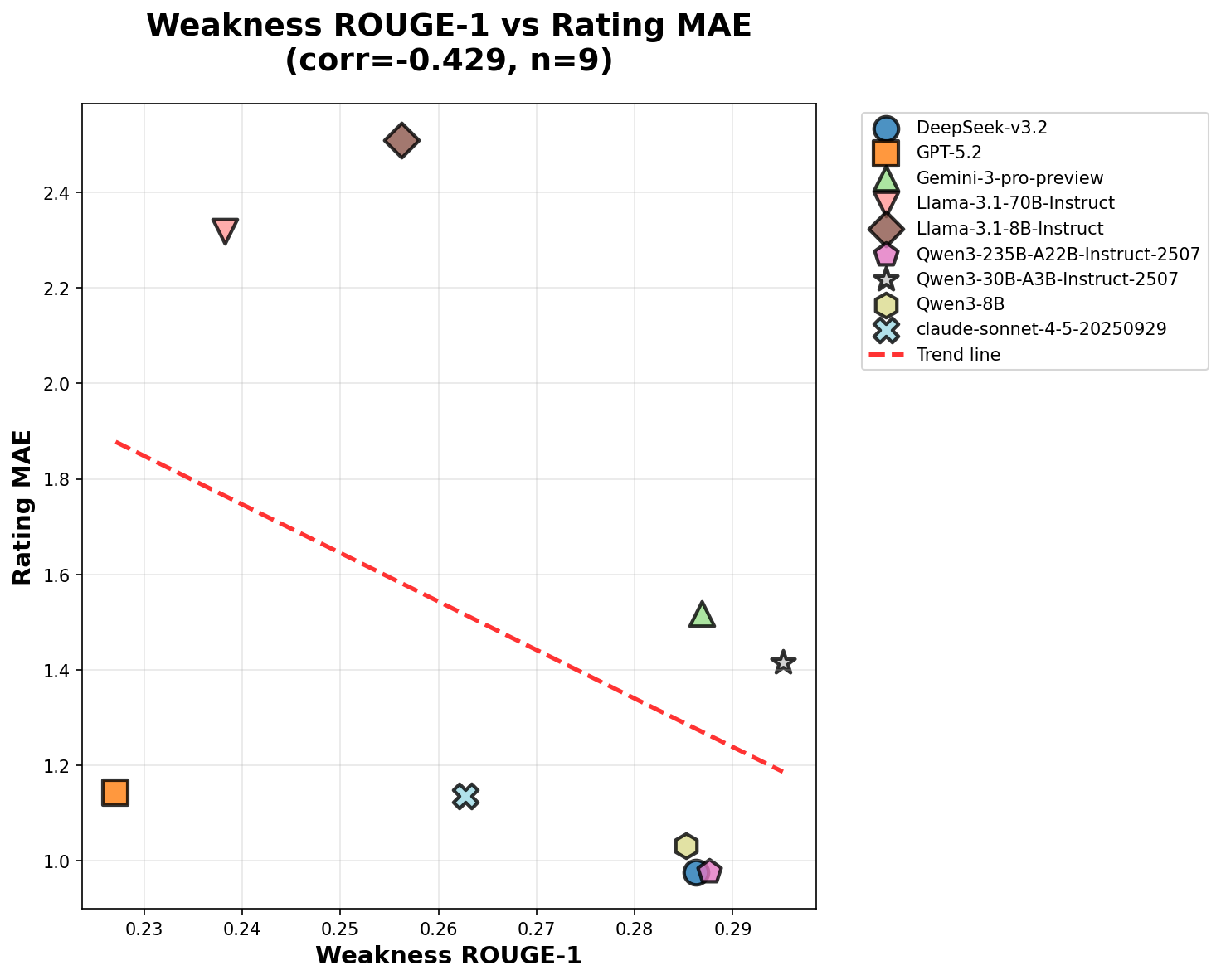}
        \caption{ROUGE-1}
        \label{fig:weakness_rouge1}
    \end{subfigure}
    \hfill
    \begin{subfigure}[t]{0.48\textwidth}
        \centering
        \includegraphics[width=\linewidth]{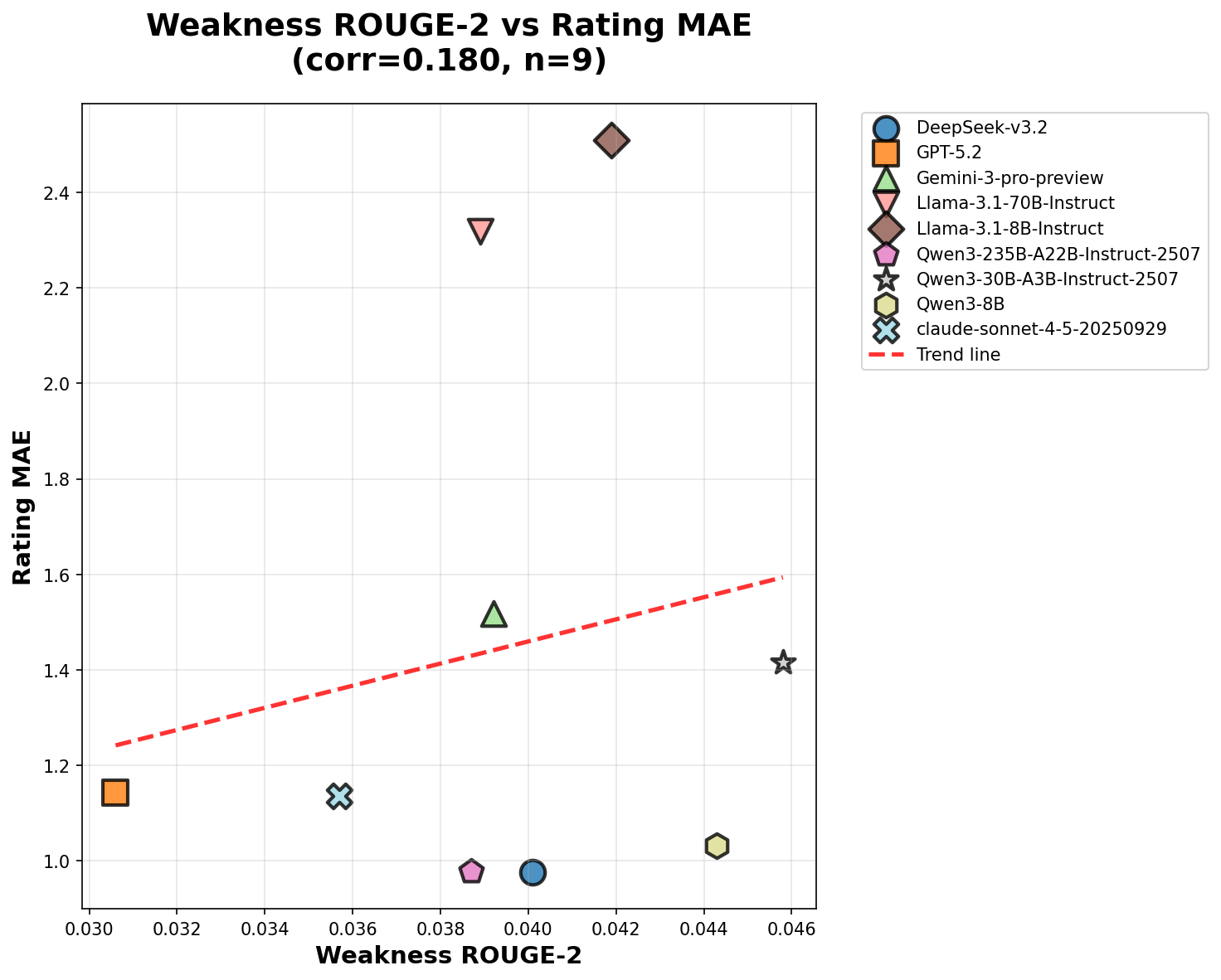}
        \caption{ROUGE-2}
        \label{fig:weakness_rouge2}
    \end{subfigure}
    
    \begin{subfigure}[t]{0.48\textwidth}
        \centering
        \includegraphics[width=\linewidth]{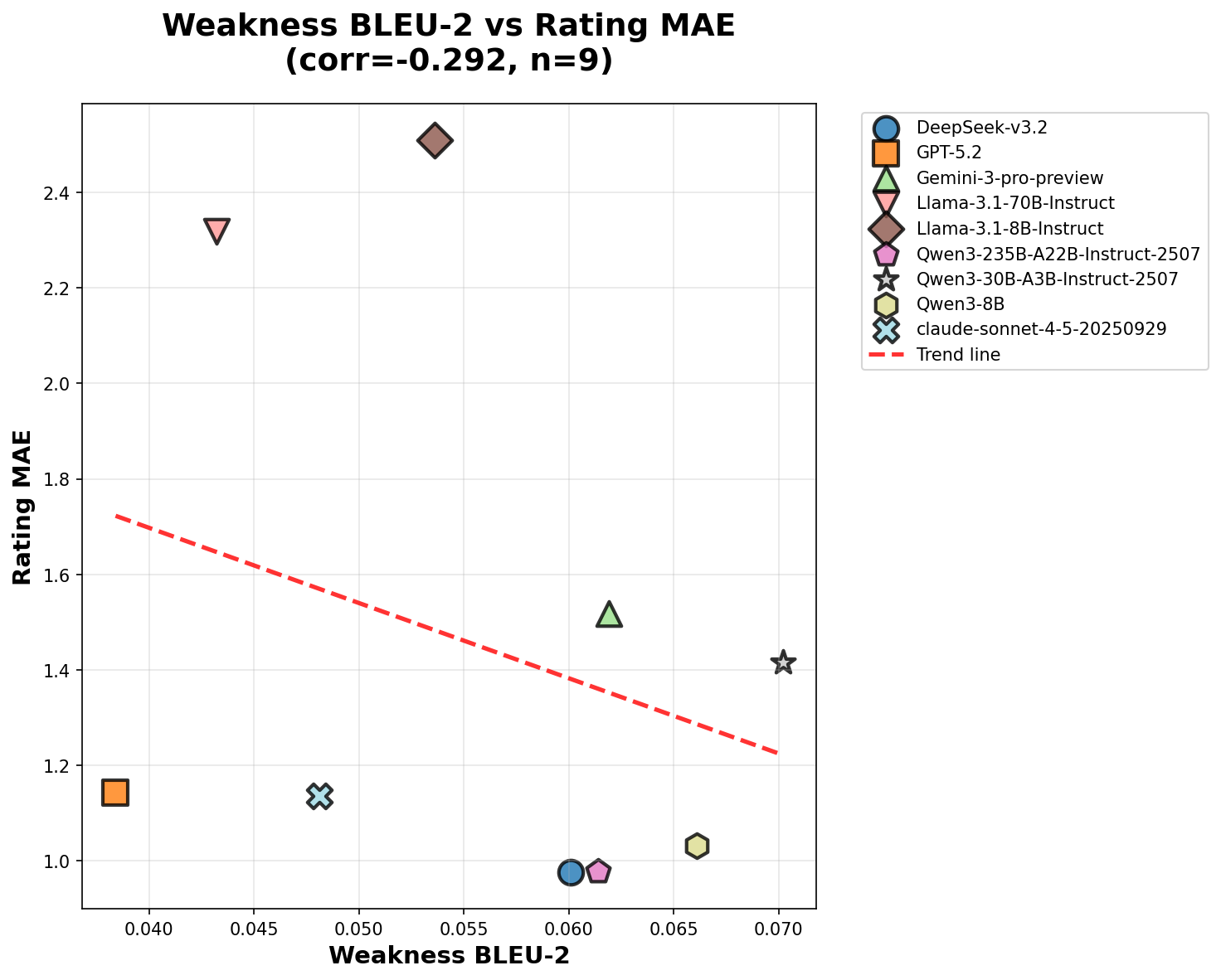}
        \caption{BLEU-2}
        \label{fig:weakness_bleu2}
    \end{subfigure}
    \hfill
    \begin{subfigure}[t]{0.48\textwidth}
        \centering
        \includegraphics[width=\linewidth]{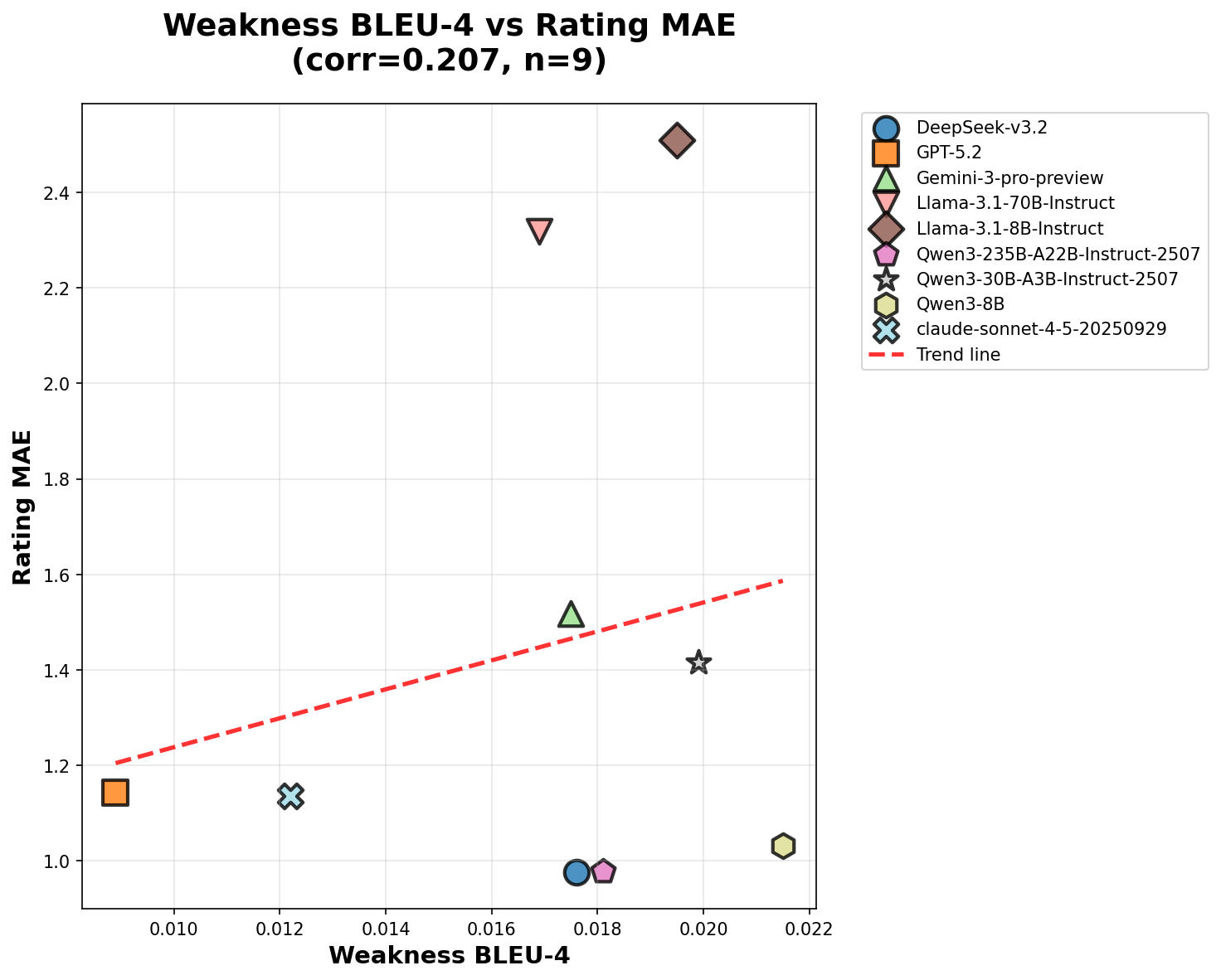}
        \caption{BLEU-4}
        \label{fig:weakness_bleu4}
    \end{subfigure}
    
    \begin{subfigure}[t]{0.48\textwidth}
        \centering
        \includegraphics[width=\linewidth]{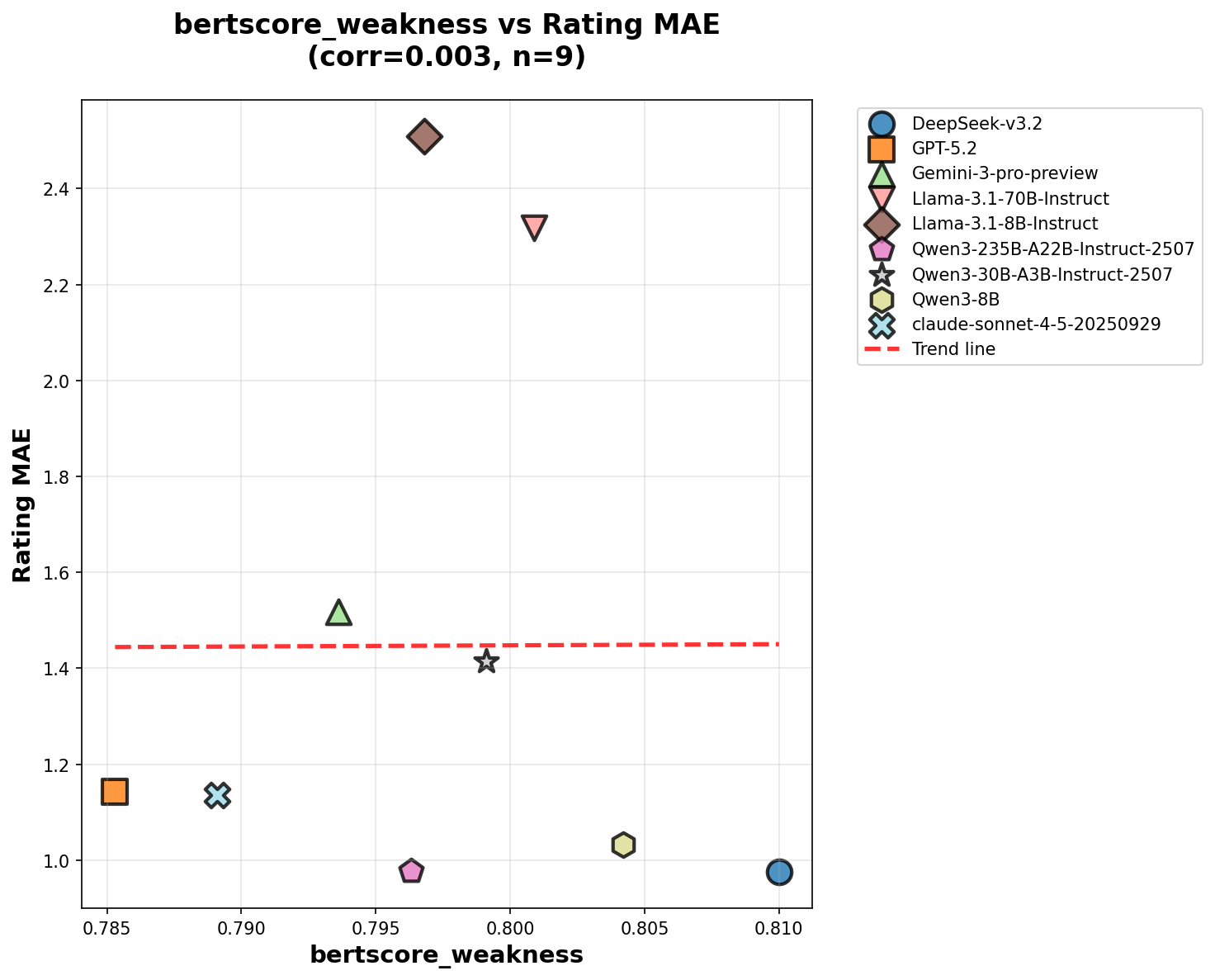}
        \caption{BERTScore}
        \label{fig:bertscore_weakness}
    \end{subfigure}
    
    \caption{Rating vs. MAE analysis for the Weakness field across different evaluation metrics.}
    \label{fig:weakness_rating_mae}
\end{figure*}

\begin{figure*}[htbp]
    \centering
    \begin{subfigure}[t]{0.48\textwidth}
        \centering
        \includegraphics[width=\linewidth]{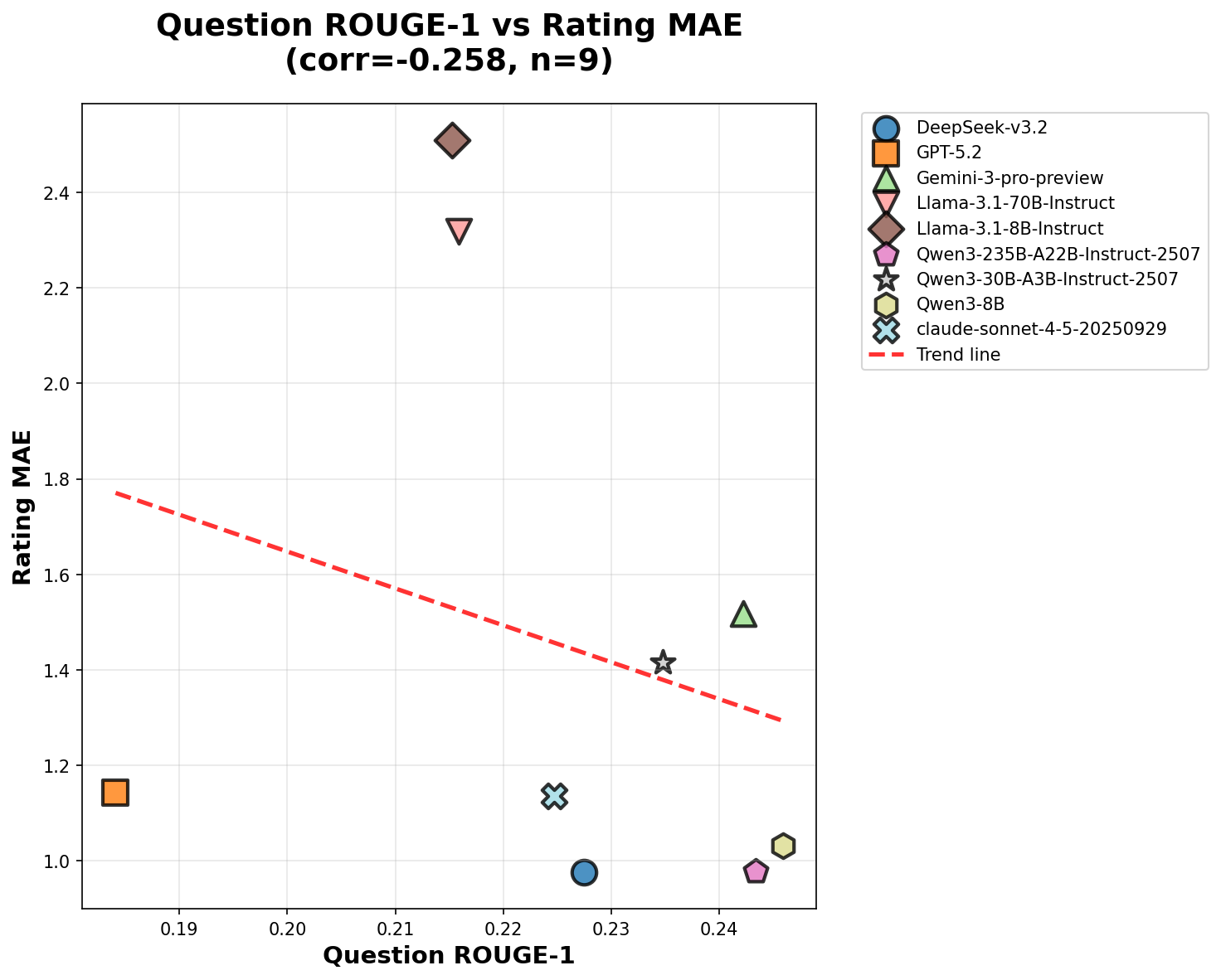}
        \caption{ROUGE-1}
        \label{fig:question_rouge1}
    \end{subfigure}
    \hfill
    \begin{subfigure}[t]{0.48\textwidth}
        \centering
        \includegraphics[width=\linewidth]{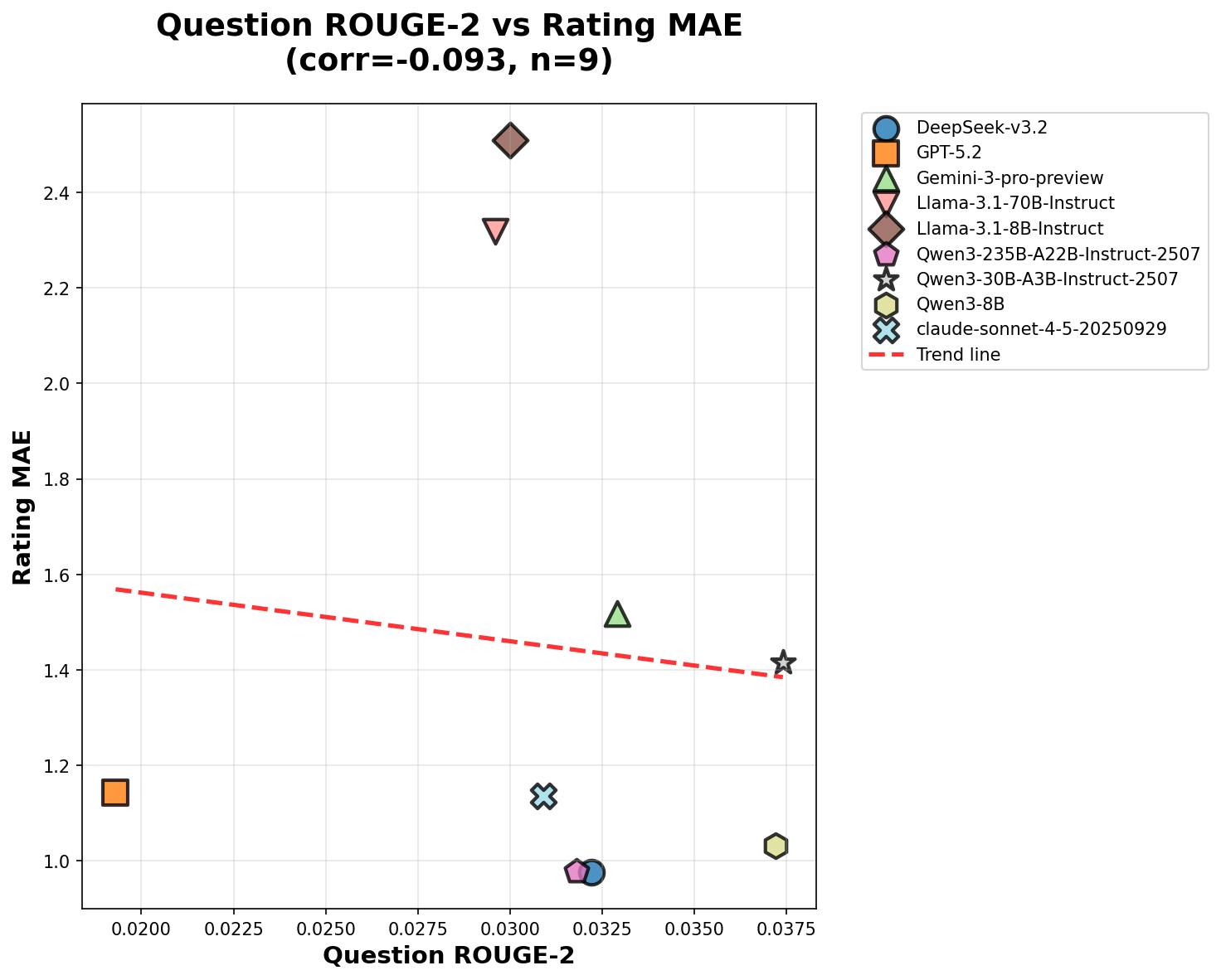}
        \caption{ROUGE-2}
        \label{fig:question_rouge2}
    \end{subfigure}
    
    \begin{subfigure}[t]{0.48\textwidth}
        \centering
        \includegraphics[width=\linewidth]{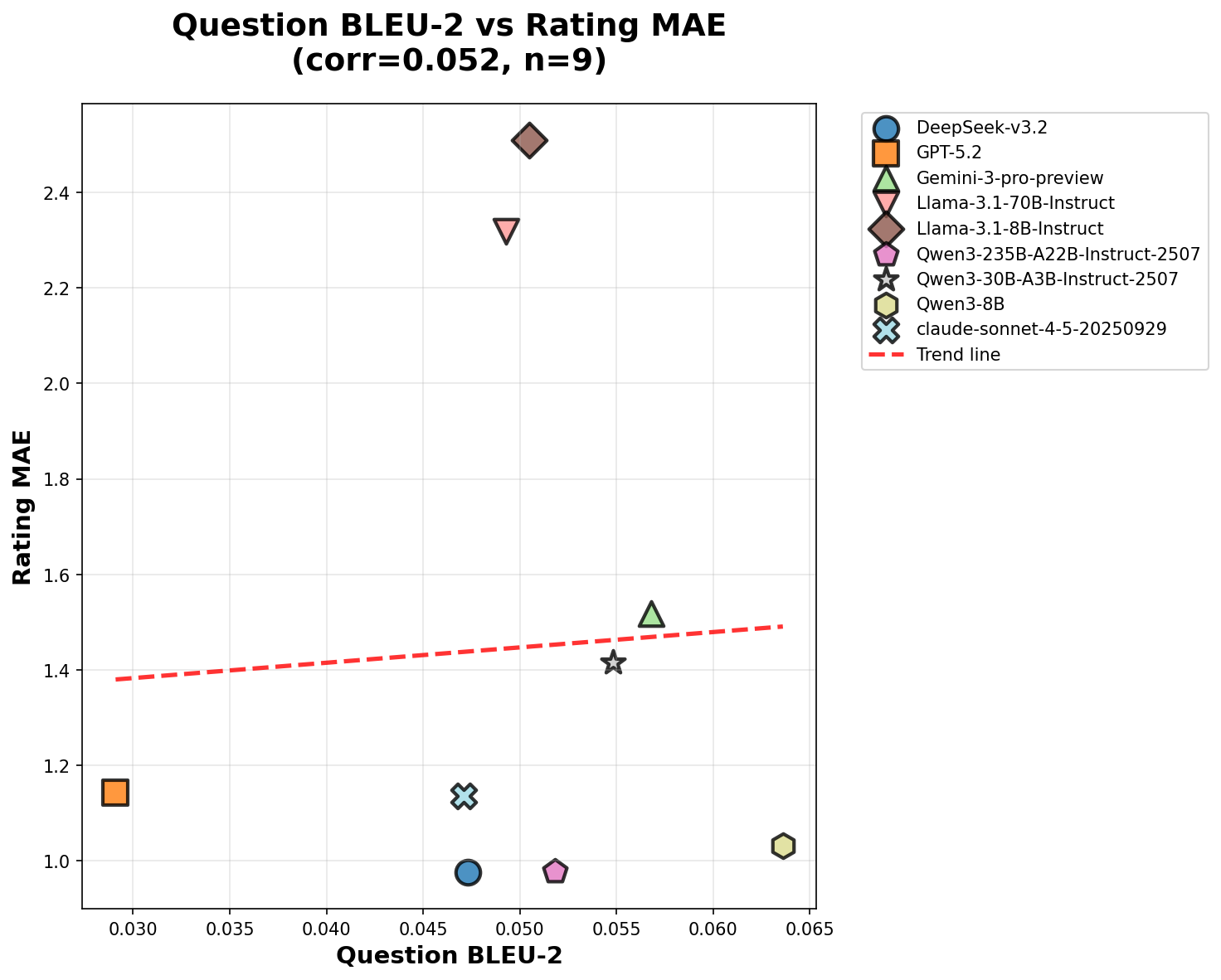}
        \caption{BLEU-2}
        \label{fig:question_bleu2}
    \end{subfigure}
    \hfill
    \begin{subfigure}[t]{0.48\textwidth}
        \centering
        \includegraphics[width=\linewidth]{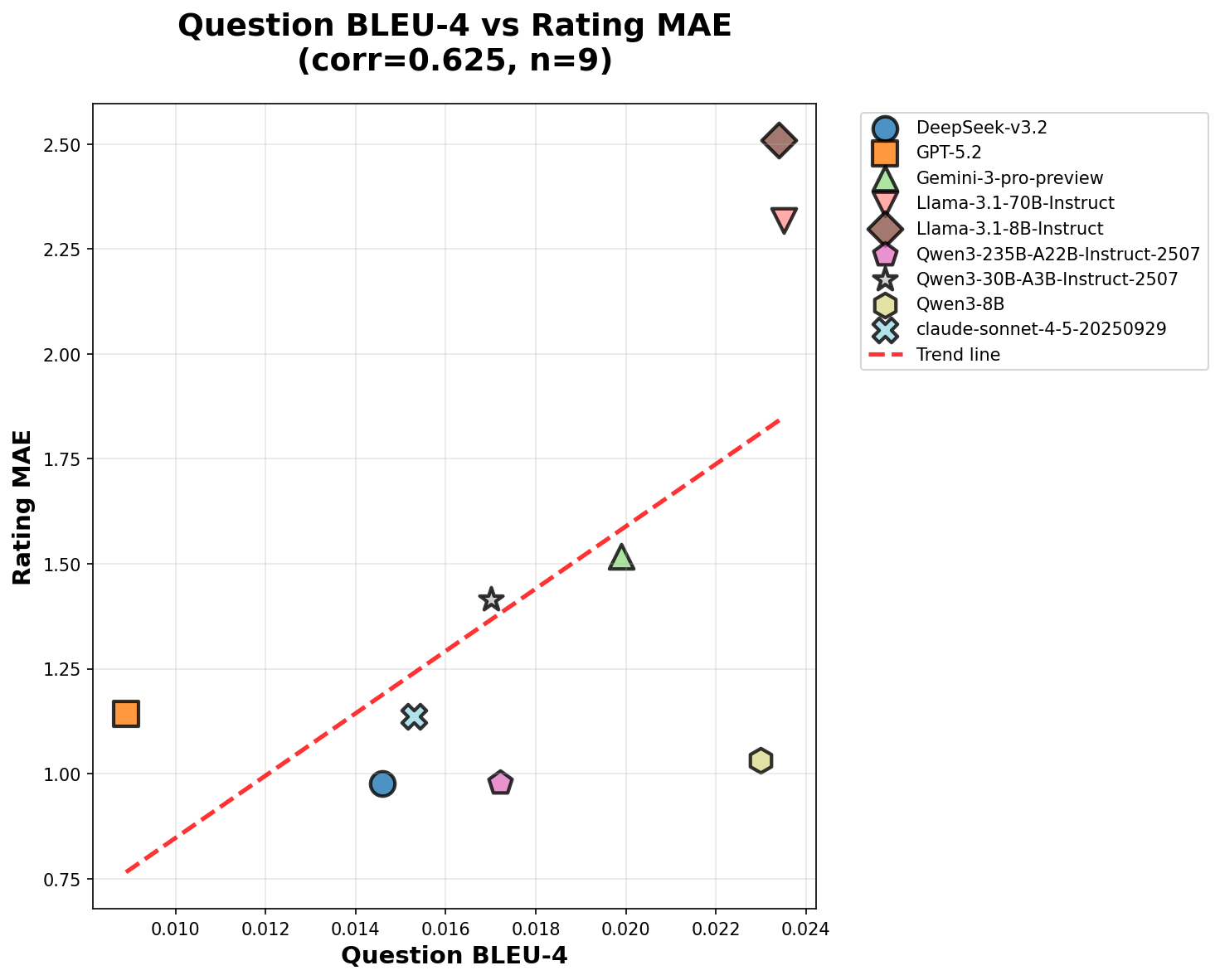}
        \caption{BLEU-4}
        \label{fig:question_bleu4}
    \end{subfigure}
    
    \begin{subfigure}[t]{0.48\textwidth}
        \centering
        \includegraphics[width=\linewidth]{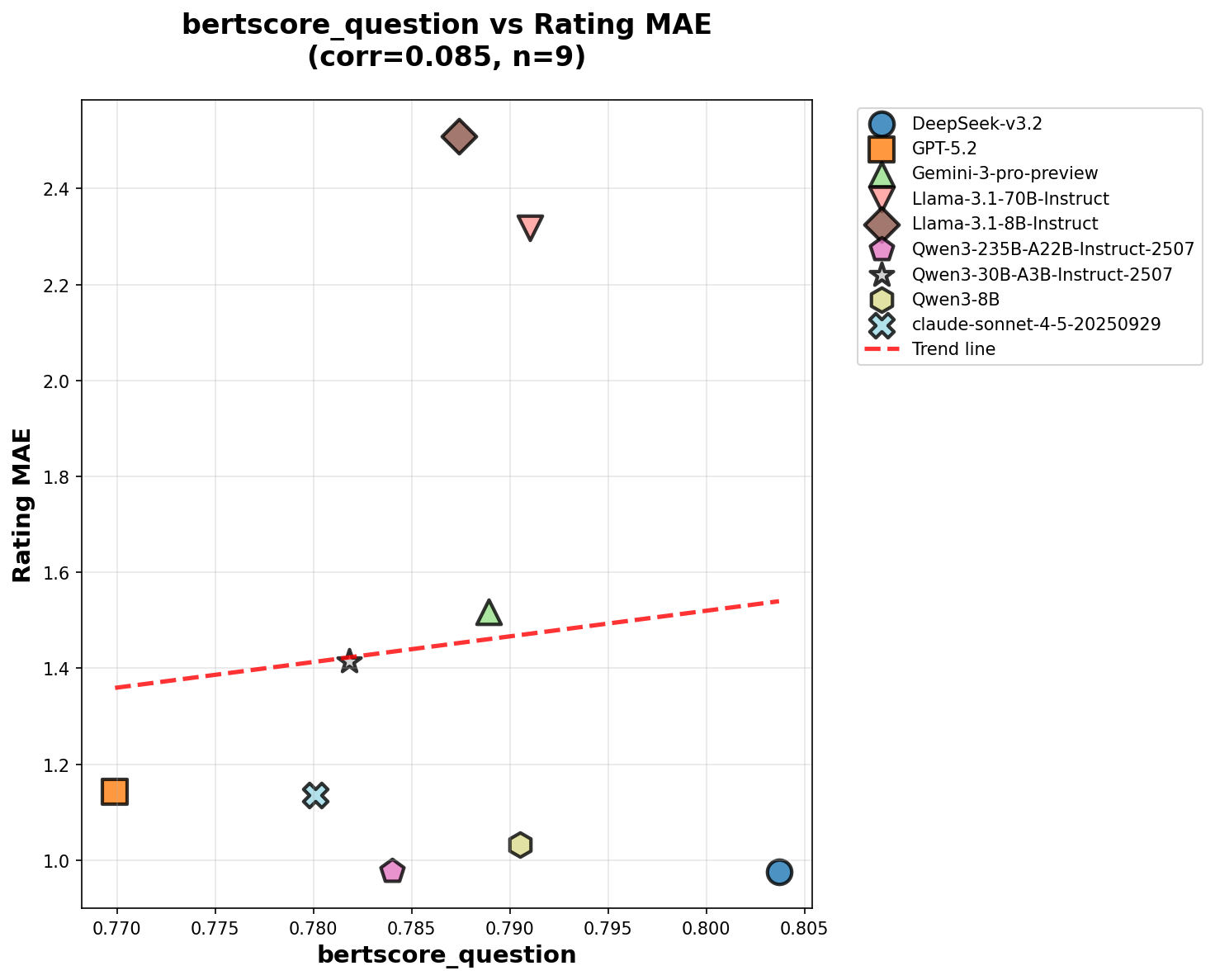}
        \caption{BERTScore}
        \label{fig:bertscore_question}
    \end{subfigure}
    
    \caption{Rating vs. MAE analysis for the Question field across different evaluation metrics.}
    \label{fig:question_rating_mae}
\end{figure*}


\onecolumn
\newpage
\end{document}